\documentclass[mksc,nonblindrev]{informs3}

\usepackage{amssymb}
\usepackage{amsmath}
\usepackage{amsfonts}
\usepackage{float}

\OneAndAHalfSpacedXI 

\usepackage{xcolor} 
\definecolor{dark-blue}{rgb}{0.1, 0.1, 0.5}
\definecolor{dark-red}{rgb}{0.6,0.15,0.15}

\newcommand\boldred[1]{\textcolor{dark-red}{\textbf{#1}}}
\usepackage{booktabs}       
\usepackage{multirow}
\usepackage{arydshln} 
\usepackage[figuresright]{rotating}
\usepackage{tabularx}

\usepackage{natbib}
 \bibpunct[, ]{(}{)}{,}{a}{}{,}%
 \def\bibfont{\small}%

 \usepackage{hyperref} 
 \hypersetup{urlcolor=dark-blue, colorlinks=true, citecolor=dark-blue, linkcolor= dark-red}
\usepackage{caption}
 \usepackage{subcaption}

\TheoremsNumberedThrough     

\EquationsNumberedThrough    

\begin{document}


\RUNAUTHOR{Kumar, Eckles \& Aral}

\RUNTITLE{Scalable bundling via dense product embeddings}

\TITLE{Scalable bundling via dense product embeddings}

\ARTICLEAUTHORS{%
	\AUTHOR{Madhav Kumar, Dean Eckles, Sinan Aral}
	\AFF{Massachusetts Institute of Technology,\\ \EMAIL{\{madhavk, eckles, sinan\}@mit.edu}} 
} 

\ABSTRACT{%
Bundling, the practice of jointly selling two or more products at a discount, is a widely used strategy in industry and a well examined concept in academia. Historically, the focus has been on theoretical studies in the context of monopolistic firms and assumed product relationships, e.g., complementarity in usage. We develop a new machine-learning-driven methodology for designing bundles in a large-scale, cross-category retail setting. We leverage historical purchases and consideration sets created from clickstream data to generate dense continuous representations of products called embeddings. We then put minimal structure on these embeddings and develop heuristics for complementarity and substitutability among products. Subsequently, we use the heuristics to create multiple bundles for each product and test their performance using a field experiment with a large retailer. We combine the results from the experiment with product embeddings using a hierarchical model that maps bundle features to their purchase likelihood, as measured by the add-to-cart rate. We find that our embeddings-based heuristics are strong predictors of bundle success, robust across product categories, and generalize well to the retailer's entire assortment.
}%



\maketitle

%



\section{Introduction} \label{s:intro}

Bundling is a widespread product and promotion strategy used in a variety of settings such as fast food (meal + drinks), telecommunications (voice + data plan), cable (tv + broadband) and insurance (car + home insurance). Given its pervasiveness, it has received considerable attention with over six decades of research analyzing conditions under which it is profitable, the benefits of different bundling strategies, and its welfare consequences. However, in spite of the vast literature, there is little empirical guidance for retailers on how to create good promotional bundles. For example, consider a medium sized online retailer with an inventory of 100,000 products across multiple categories. Which two products should the retailer use to form discount bundles? There are ${10^5 \choose 2} \approx 50$ million combinations. Conditional on selecting a candidate product, there are 99,999 options to choose from to make a bundle. Is there a principled way that the managers can use to select products to form many bundles?

In this study, we offer a new perspective to the bundle design process which leverages historical consumer purchases and browsing sessions. We use them to generate latent dense vector representations of products in such a way that proximity of two products in this latent space is indicative of ``similarity'' among those products. A key insight in our method is the distinction between the representation of product purchases and representation of consideration sets, where consideration sets include the products that were viewed together but \textit{not} purchased together. We posit that products that are frequently bought together tend to be more complementary whereas products that are frequently viewed but \textit{not} purchased together tend to be more substitutable. Then, depending on whether the products were frequently co-purchased or co-viewed (but \textit{not} purchased), the degree similarity in the latent space is suggestive of complementarity or substitutability respectively. We put minimal structure on this latent-space-based contextual similarity to generate product bundles. We then learn consumers' preferences over these suggested bundles using a field experiment with a large U.S.-based online retailer. Finally, we generalize our findings to the entire product assortment of the retailer by modeling the bundle success likelihood, as measured by the bundle add-to-cart rate, as a function of the product embeddings using hierarchical logistic regression. 

Much of the earlier work on bundling was from economists seeking conditions under which a monopolist might choose to sell its products as independent components vs. pure bundles vs. mixed bundles \citep{AY1976, Schmal1982, Venkatesh2003}. These earlier papers were motivated by considerations of price discrimination and hinged on analytical models that rely on pre-specified product complementarity and substitutability, or an ex-ante well-defined underlying relationship between the products.  For instance, \cite{AY1976} conduct their analysis by assuming that the products have independent demands, \cite{Venkatesh2003} provide conditions for products that are assumed to be either complements or substitutes, and, more recently, \cite{Kumar2013} develop their model in the context of video-games and consider obvious complements in usage --- consoles and games. Furthermore, most studies work with the idea that a single firm is producing the goods, bundling them together, and then selling them at the discounted price. However, a more realistic picture --- and the one we consider in this study --- is one of a downstream retailer bundling products from different firms. 

Our work enhances the existing literature on bundling in economics and marketing in three ways. First, instead of considering pre-defined relationships among products, we generate continuous metrics that are heuristics for the degree of complementarity and substitutability based on historical consumer purchases and consideration sets. A major strength of our approach is that we can learn relationships between two products which may have never been co-purchased or co-viewed together but still be strongly related to each other. This permits us to develop an effective bundle design strategy in a large-scale cross-category retail setting where co-purchases are sparse, a relatively unexplored context in bundling studies. Second, we test the effectiveness of our methodology by running a field-experiment with a large online retailer in the US, providing empirical color to a largely theoretical literature. Third, we explore the idea of generating bundles from imperfect substitutes to tap into the variety-seeking behavior of consumers, which we call \textit{variety} bundles.

This paper also provides implementable insights for managers. Identifying the best bundles for a retailer with an assortment of a 100,000 products involves considering an action space with millions of potential bundles, a combinatorially challenging task. Our methodology allows us to filter this action space in a principled data-driven way using machine-learning based heuristics, providing substantial efficiency gains while accounting for consumer preferences. For example, some of the bundles created by category managers include different volumes of the Harry Potter book series, branded sports team gear (hand towel and bath towel), and same-brand shampoo and conditioner. Our approach adds several types of bundles to this set: cross-category complements with a centerfold table and a single-door compact refrigerator (furniture + electrical appliances), cross-(sub)category complements with mouthwash and deodorant, within-category complements involving laundry detergent and stain remover, and variety snacks with potato chips and animal crackers. Moreover, since we run the experiment at a product level, we are able to flexibly generate micro-level insights, e.g., which brands make good bundles, as well as high-level insights such as which categories make good bundles.

\begin{figure}[h]
	\centering
	\includegraphics[scale=0.45]{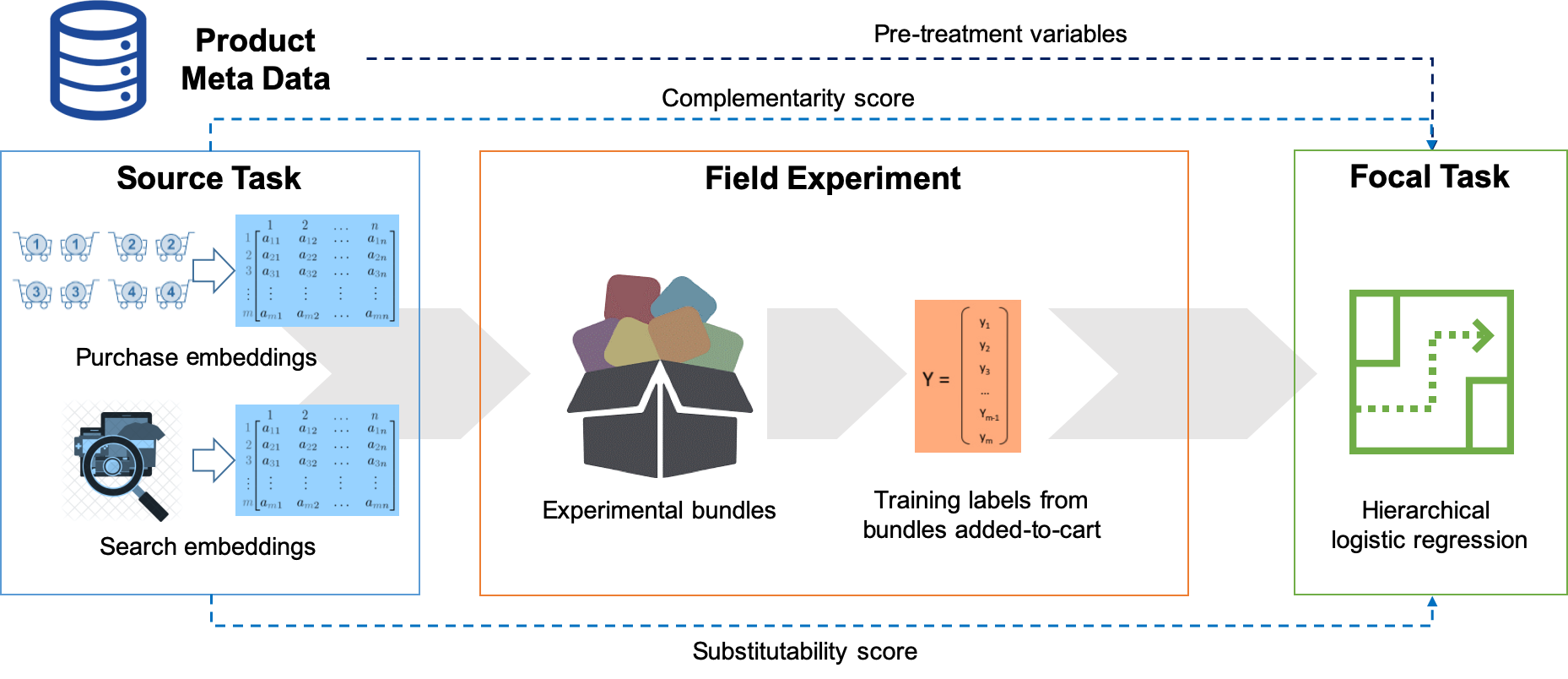}
	\caption{Transfer learning framework for creating product bundles}
	\label{f:tl_frame}
\end{figure}

Our methodology is setup as a transfer learning \citep{Ng2006,Pan2010} framework, shown in Figure~\ref{f:tl_frame}, an approach used in machine learning which involves employing the knowledge learned in one task, called the source task, to a related task, called the focal task.  The source task is typically one in which there is more data or prior knowledge available, whereas the focal task has much less data to perform standalone robust statistical analysis. Our focal task is to create good promotional bundles. However, we do not have a principled way of selecting which products would make good bundles. Our only prior is co-purchase frequency, which by itself is too sparse for generalizable findings. What we do have is a high-volume of historical product purchases and consideration sets. Using them, we create a source task and use unsupervised learning to learn product relationships and generate heuristics for complementarity and substitutability. We validate the effectiveness of these heuristics for generating good bundles through a field experiment and then use the results from the experiment as training data for our focal task. Our approach gradually moves from an unsupervised machine learning model to a supervised hierarchical model, leveraging the respective strengths of methods for the underlying tasks. As as aside, we also hope that this study serves as a useful guidepost for researchers trying to effectively use machine learning methods in conjunction with econometrics.

The paper is structured as follows. We describe the previous work in the next section. Our model is described in Section \ref{s:model}, which is followed by a description of the data and the product embeddings. The details of the field experiment are presented in Section \ref{s:exp} and the insights from predictive modeling are shown in Section \ref{s:predict}. Section \ref{s:disc} concludes.

\section{Related work} \label{s:lit}

Our study draws inspiration from two distinct strands of literature: the bundling literature from economics, marketing and operations research (OR), and the machine learning work from natural language processing (NLP). 

\subsection{Economics, marketing, and OR}

The idea of bundling is described visually and intuitively by \cite{AY1976}, who show that bundling is profitable in the case of goods with independent demands because firms are able to sort consumers based on reservation prices, which in turn allows them to extract a larger portion of consumer surplus. \cite{Lewbel1985} extends \cite{AY1976} to include interdependent demand and shows that it may not always be profitable to bundle complements. He goes further to explain that a multi-product monopolist may actually find it profitable to bundle imperfect substitutes by using mixed bundling. \cite{Schmal1982} provides the conditions for pure bundling (selling products only as a bundle) and mixed bundling (selling products independently and as a bundle) in the case of a single-product monopolist and a competitively sold product as a function of reservation prices, production costs, and market structure. He finds the mixed bundling can be profitable when there is negative correlation between the reservations prices of the two goods. In a separate paper, \cite{Schmal1984} attempts to generalize \cite{AY1976}'s findings by assuming that the distribution of reservation prices is bivariate normal. With the Gaussian tool (and its caveats) in hand, he finds that with symmetric reservation price distribution, pure bundling is better than unbundled sales. \cite{Salinger1995} uses the concept of negatively and positively correlated demand to show the welfare consequences and profitability of bundling graphically. His primary focus though is on the relationship between demand for the bundle and the demand for individual components of the bundle.\footnote{\cite{Stremersch2002} is an excellent introductory guide to the concept of bundling from a marketing perspective and also provides generic conditions under which one form of bundling may be preferred over another. } 

\cite{Venkatesh2003} consider a multi-product monopolist and build an analytical model based on contingent values to identify conditions when bundling complements, substitutes, and independently valued products works. \cite{Venkatesh1993} take a probabilistic approach to find conditions suitable for pure components, pure bundling or mixed bundling. They include time availability along with reservation price as another dimension for consumer evaluation and for each bundling strategy, calculate the optimal prices, profits, and market share. \cite{Ansari1996} extend their work by allowing the number of components to be endogenously determined. They apply their model to the non-profit space by studying survey data on classical music events.

While the setting in most economics papers has been a monopolist firm of some kind, a few papers in marketing have focused on the retail perspective as well. \cite{Mulhern1991} consider retail pricing and develop a theoretical framework based on demand interdependencies. They consider the case of multi-product pricing based on the notion that promotion pricing of related products is equivalent to implicit price bundling. Relationships are defined in such a way that products within a product line are substitutes and those across product-lines can be either substitutes, complements, or independent products. They then show how retailers can use price promotions using store-level weekly scanner data on cake mix and frosting. \cite{Rao2003} build a multi-category choice model for bundles based on the attributes of the products. They estimate their parameters by pre-defining a set of physical features and attributes for personal computers. Although, their choice model does account for cross-category bundles and hence, heterogeneous components, their definition of categories is fairly narrow, with all products being complements in usage. \cite{Bhargava2012} studies the impact of a merchant bundling products from different firms. He author builds an analytical model to find conditions under which pure bundling and pure components are profitable. He further shows that bundling may not be profitable due to vertical and horizontal channel conflicts unless the firms can co-ordinate on prices. 

Among empirical works, \cite{Yang2006} use association rules to create bundles of books based on shopping-cart data and browsing data. The find that these bundles are better, as measured by total number books bought, than the bundles based solely on order data or solely using browsing data. Although, \cite{Yang2006}'s idea and our idea are similar in spirit, i.e., both use browsing and purchase data to generate bundles, our scopes are widely different - books vs. cross-category retail. Nevertheless, a key takeaway from their study is that they found incremental value in using browsing data in addition to purchase data in forming bundles, an idea that we leverage too. 

\cite{Jiang2011} also study bundling in the context of an online retailer selling books and use non-linear mixed integer programming to recommend the next best product given what it is currently in the basket. They do numerical studies to show that their method leads to more customers purchasing discounted bundles as well as improved profits for the retailer. More recently, \cite{Kumar2013} empirically test some of bundling theories described above in the context of hand-held video games. The investigate the options of pure bundling vs. mixed bundling along with the dynamic effects of bundling for durable complementary products. They find that mixed bundling leads to higher revenues than pure bundling or pure components. Bundling also causes consumers to behave strategically in that they lead certain segments of consumers to advance their purchases and certain segments to enter the market which they might not have in the absence of bundling. 

To summarize, previous research has carefully examined the efficacy of different bundling strategies in a variety of settings with a multitude of tools such as graphical analysis, analytical frameworks, probabilistic, and structural models, survey-based empirical exercises, and modeling historical purchase data. Conclusions, though numerous, are contingent on the assumptions the researchers have made. Depending upon the context, researchers have found bundles of complementary, substitutes, and independent products to be profitable. With lessons from these papers as strong a foundation, our paper offers a new perspective to an old problem. We do not attempt to fill any ``gap'' in the literature but rather deliver a novel prescriptive methodology that is rooted in data, is empirically validated using a field experiment, and is practically implementable by managers and retailers.

\subsection{Machine learning}

We leverage techniques developed for analyzing large unstructured data, especially text data, in computer science called neural embeddings. The basic idea behind these methods is to convert textual representations of words to numerical vectorized representations which are dense and continuous \citep{Mikolov:2013a, Mikolov:2013b}. Frequent co-occurrences of words within and across documents are indicative of semantic and syntactic relationships between them. Neural embeddings translate co-occurrence patterns in text into a latent space of a pre-specified dimension, say $\mathcal{D}$, where proximity in the latent space implies semantic similarity. We don't discuss the core literature here but rather focus on the applications of it in our domain. The interested reader can look at \cite{Mikolov:2013a, Mikolov:2013b}. 

The concept of latent embeddings has been applied to diverse settings such as reviews, neural activity, movie ratings, product recommendations, and even market baskets \citep{Rudolph:2016, Rudolph:2017, Shopper2017, item2vec-2016, Hauser2018}. The core data framework in these papers is similar to ours, in which there is unstructured data of a sequence of objects generated through repeated actions of an agent. Those actions could be rating different movies by a viewer \citep{Rudolph:2016}, or listening to songs \citep{item2vec-2016}, or purchasing multiple products together \citep{Rudolph:2016, Shopper2017}. Elaborating with the example of purchase baskets, the fundamental action in these scenarios is binary, i.e., purchasing a product or not, which can be thought of a word being present or not. A sequence of such binary activities (words) then makes up a sentence, which in our example is a product basket, and multiple product baskets constitute to form the entire dataset of purchases made by different consumers over time. Interpreting these data structures as the outcome of an underlying text generation process allows us to leverage the algorithms from natural language processing and bring them to the marketing domain. Our modifications to the original algorithm are described in Section \ref{s:model}. 

Among the papers cited above, \cite{Shopper2017} is closest to our work. They look at product baskets to build a model of consumer choice, eventually generating latent representations of products that can then be used to identify economic relationships among products such as complementarity and substitutability. We find their work insightful since our setting is quite similar --- we also inspect product baskets to generate dense latent representations of products and then eventually use them to learn product relationships. However, there are three important distinctions. First, they only consider products that were purchased together and use the embeddings from the purchase space to determine complements and substitutes. We, on the other hand, use clickstream data that allows us to identify consideration sets before purchases and define a heuristic of substitutability through products that are viewed together but not purchased together. Second, our ultimate objective is different from theirs. They propose a novel model in the utility-choice framework; we are in-effect taking the utility-choice framework as given and using our version of that framework to design retail product bundles. Third, though less important, is that our model training approaches are different. Their approach is based on variational inference while ours is based on a shallow neural network.

\section{Model} \label{s:model}

Inferring product relationships from consumer choice has largely been the bastion of economists studying mirco-econometric discrete choice models of consumer demand in which a consumer typically chooses one product out an assortment of within-category options. The within-category part is important since it constrains the consumer's choice set to close (but potentially imperfect) substitutes, rendering cross-category comparison extremely difficult. Additionally, most of these models are limited in the number of products they can handle and also in the number of transactions, though the latter concern has been ameliorated with rise in computational power. Furthermore, previous models only allow us to use features of products that are easily observable and quantifiable such as brand, price, and size. However, consumers make choices based on many factors such as the product description, packaging, and reviews, all of which are not only difficult to quantify and also unintuitive to compare across categories of products. 

Our approach loosens the grip of all these constraints by (1) considering multi-product choices in the same shopping session, (2) leveraging cross-category purchase baskets and consideration sets, (3) ensuring scalability in number of products and the number of shopping sessions, and (4) imposing minimal structure on product characteristics. For example, in our setting of online retail with $35,000$ products across hundreds of categories, inferring relationships between products through cross-price elasticity is not feasible. Co-purchases at the product level are too sparse to generate reliable estimates. Over 90\% of the product pairs have never been purchased together. To analyze this sparse high-dimensional data efficiently, we adapt methods from the machine learning literature and customize them to suit our case. Our model condenses a large set of information about each product into dense continuous vector representations, which facilitate easy comparison of products across categories. Moreover, our method is also useful when considering categories of products with thin purchase histories, an area which is particularly difficult for structural choice models, allowing us to infer relationships even among products with few purchases. 

Our model belongs to the general class of vector space models used to embed where discrete tokens can be represented as continuous vectors in a latent space, such that tokens that are similar to each other are mapped to points that are closer in the space. Popular examples of vector space models include \textit{tf-idf}, and the relatively newer, \textit{word2vec} \citep{Mikolov:2013a,Mikolov:2013b}. Though both the examples above rely on the distributional hypothesis, models such as \textit{tf-idf} are commonly referred to as count-based methods and are based on coarse statistics of co-occurrences of words in a text corpus, where models such as word2vec are based on prediction methods \citep{Baroni2014}. While a count-based model, such as an n-gram  \textit{tf-idf}, is simpler to understand, estimating the parameters of becomes increasingly complex as n increases $\big( \mathcal{O}(\mathcal{\vert V \vert}^n) \big)$, where $\mathcal{\vert V \vert}$ is the size of the vocabulary. Count-based methods also cause problems when they face unforeseen n-grams and require smoothing to deal with them. 

Neural probabilistic language models, like word2vec, deal with both these concerns by changing the objective function from modeling the likelihood of the corpus to predicting the probability of a word given its surrounding words. This not only condenses the representation of each word to a much lower dimension as compared to the size of the vocabulary but also removes the need for smoothing to generate probabilities estimates for new word sequences. It is important to note here that while neural models also rely on heavily on co-occurrences, they go much beyond the simple notion of co-occurrence to help us learn about word pairs that may not have been frequently observed together in the past. In our context (as we will explain below), this implies that we can learn about product pairs that may have had historically low co-purchases but could still be strongly related to each other. In the following sub-sections, we first provide an intuition behind our model and then formalize it.

\subsection{Intuition}

Our model is a customized version of a widely used shallow learning technique from the machine learning literature used to analyze discrete, sparse data \citep{Mikolov:2013a, Mikolov:2013b}. It has been fairly popularized in recent years due to its application in analyzing text. In the language processing field, the intuition behind this method is simple --- words that occur frequently together in the same context are likely to have a semantic, and syntactic, relationship with each other. For instance, consider the following sentences:
\begin{align*}
& \text{Esha has \boldred{milk}, \boldred{cereal}, and \boldred{coffee} for breakfast} \\
& \text{The tragedy is that she pours her \boldred{milk} before the \boldred{cereal}} \\
& \text{She also has \boldred{coffee} with \boldred{milk} in the evening} \\
& \text{She prefers \boldred{coffee} with a little bit of \boldred{sugar}}
\end{align*}
In these sentences, milk and cereal appear together frequently (relatively speaking) and that milk and coffee also appear together frequently. Our understanding of language plus banal observation of the world tell us that milk and cereal are ``related'' and that milk and coffee are also ``related''. Essentially, these are the kinds of associations that we attempt to capture with our model, albeit with some refinements. 

Translating the language from text documents to retail products, we exploit the notion of product baskets, i.e., we take products purchased together by consumers, and think of them as text sentences. The underlying idea is that products that appear frequently together in multiple baskets have a relationship that is beyond mere random co-occurrence. To make this idea clear, consider Esha's consumption basket as shown below. It reproduces the sentences from above with everything but the products consumed removed. For the sake of exposition, we also add variants of the products consumed. The baskets then look like:
\begin{align*}
& b_1 : \text{low fat milk, crunchy cereal, dark roast coffee} \\
& b_2: \text{low fat milk, dark roast coffee} \\
& b_3: \text{dark roast coffee, raw sugar}
\end{align*}
These baskets are perfectly valid sentences for our algorithm to process with each product being a word and each sentence being a combination of these products. We can then build a model similar to the one used in NLP to learn relationships among products, with two important caveats: (1) the order of the products in our basket does not matter, and (2) our model needs to consider only two products at a time since we are building bundles with only two-component products. We thus transform each basket to a two-product combination with all possible permutations. This gives us the following baskets:
\begin{align*}
& b_{11} : \text{low fat milk, crunchy cereal} \\
& b_{12} : \text{crunchy cereal, dark roast coffee} \\
& b_{13} : \text{low fat milk, dark roast coffee} \\
& b_{21}: \text{low fat milk, dark roast coffee}\\
& b_{31}: \text{dark roast coffee, raw sugar}
\end{align*}

This transformation effectively converts our unstructured data to a simple classification problem where all the instances above form positive cases. To operationalize this model, we need two more inputs: (1) negative cases for the model to distinguish between products purchased together and products not purchased together, and (2) an optimization algorithm to learn the parameters. One can simply sample negative cases by considering pairs of products that do not occur in the same baskets, but are present in the inventory \citep{Mikolov:2013b}. Now, with both positively labeled samples and negatively labeled samples, we can run our favorite classification algorithm to train the parameters. Of course this is an overly-simplified stylized example. We present a more formal treatment of the underlying process and the model in the next sub-section. 

To complete the picture, along with products purchased together, we also consider products from users' consideration sets. Analogous to purchase baskets, we form search baskets, i.e., products which were viewed but \textit{not} purchased together, break them into pairs of two products to form positive cases, and likewise generate negative cases. We explain the motivation behind our use of purchase and search baskets later in the text. 

Lastly, with recent advances in machine learning methods and computational infrastructure there are now multiple ways to train these models (e.g., word2vec\footnote{\url{https://radimrehurek.com/gensim/models/word2vec.html}}, glove\footnote{\url{https://nlp.stanford.edu/projects/glove/}}, fasttext\footnote{\url{https://fasttext.cc/}}). We write our own version of the model in Tensorflow\footnote{\url{https://www.tensorflow.org/}}, which we describe below in the context of purchase baskets. The reasoning can be easily extended to the concept of search baskets. 

\subsection{Formal model}

Consider a retailer with a assortment $\mathcal{ V }$ of size. Suppose our representative consumer, Esha, purchases 5 products, forming the product basket $b_1$: $\{w_1, w_2, w_3, w_4, w_5 \}$. Our objective is then to predict the products $\{w_2, w_3, w_4, w_5 \}$ given the product $w_1$. Unlike natural language models, we do not consider the order of the products, bu use the entire remaining basket to be the \emph{context} for product $w_1$. Let $\mathcal{C}$ be the set of context products, such that, $\mathcal{C}(w)$ represents the set of products in the context for product $w$. With the basket above, given the product $w_1$ and its context $\mathcal{C}(w_1) = \{w_2, w_3, w_4, w_5 \} $, we want to maximize the log-likelihood of the basket,
\begin{align}
\mathcal{L}_{b_1} =  \sum_{w \in b_1} \log P \big(\mathcal{C}(w) \vert w \big).
\label{e:obj}
\end{align}

Here we introduce the concept of embeddings, the dense continuous representations we are trying to estimate. Suppose that each product in the assortment is represented by two $\textit{d}$-dimensional real-valued vectors, $v$ and $u$. The matrices $\mathbb{U}$ ($\mathcal{\vert V \vert} \times \textit{d}$) and $\mathbb{V}$ ($\textit{d} \times  \mathcal{\vert V \vert}$) are the emebdding matrices, where $u_i$ and $v'_i$ give two representations for product $w_i$. $\mathbb{V}$ is the input matrix and $\mathbb{U}$ is the output matrix. The process of predicting $\mathcal{C}(w_1)$, given $w_1$ boils down to estimating the probability $P \big(\mathcal{C}(w) \vert w \big)$ mentioned in \ref{e:obj}. Considering one element $w_2$ from $\mathcal{C}(w_1)$, we can write this probability using the logit model,
\begin{align}
P \big(w_2 \vert w_i \big) = P(u_2 \vert v'_1) =  \frac{\exp(u_{2} \cdot v'_{1})}{\sum_{k = 1 }^{\mathcal{\vert V \vert}} \exp(u_{2} \cdot v'_{1})},
\label{e:prob_exp}
\end{align}
where $u_2$ is the second row from the output embedding matrix $\mathbb{U}$ and $v'_1$ is the first column from the input embedding matrix $\mathbb{V}$. 

Generalizing expression \ref{e:prob_exp} to account for all products in the context, we can write the conditional probability term in the objective function shown in \ref{e:obj} as:

\begin{align}
P \big(\mathcal{C}(w_i) \vert w_i \big) = \prod_{w_j \in \mathcal{C}(w)} \frac{\exp(u_{w_j} \cdot v'_{w_i})}{\sum_{k = 1 }^{\mathcal{\vert V \vert}} \exp(u_{w_k} \cdot v'_{w_i})}
\end{align}

A point to note is the calculation of the denominator in the above expression. Typically, $\mathcal{\vert V \vert}$ is quite large and hence for computational efficiency we employ negative sampling as described in \citep{Mikolov:2013b} to approximate the denominator. With negative sampling, we use only select a sample of the negative examples to update at each iteration. We use a unigram distribution to sample negative examples such that more frequently occurring products across baskets are selected more likely to be chosen. Assuming we select, $N_s$ negative examples, we can write the approximate probability expression as
\begin{align}
P \big(\mathcal{C}(w_i) \vert w_i \big) = \prod_{w_j \in \mathcal{C}(w)} \frac{\exp(u_{w_j} \cdot v'_{w_i})}{\sum_{k = 1 }^{N_s} \exp(u_{w_k} \cdot v'_{w_i})}.
\end{align}
Plugging this value in the log-likelihood function to estimate the probability of each product in the context $\mathcal{C}(w_i)$ for given a target product $w_i$, we get
\begin{align}
\mathcal{L}_{b_1} =  \sum_{w \in b_1} \bigg[ \sum_{w_j \in \mathcal{C}(w_i)} \bigg( \log {\sigma(u_{w_j} \cdot v_{w_i})} + \sum_{k \ne j, k = 1}^{N_s} \log{\sigma(-u_{w_k} \cdot v_{w_i})} \bigg) \bigg],
\label{e:obj_skgm}
\end{align}
where $\sigma(x) = \frac{1}{1 + \exp(-x)}$ is the sigmoid function. 

We estimate the parameters $\mathbb{U}$ and $\mathbb{V}$ by maximizing the likelihood of all baskets in the data set. The log-likelihood for the entire data set is given in Equation \ref{e:total_ll}, where $\mathcal{B}$ is the set of all product baskets observed in the data,
\begin{align}
\mathcal{L}_{\mathcal{B}} =  \sum_{b \in \mathcal{B}} \sum_{w \in b} \bigg[ \sum_{w_j \in \mathcal{C}(w_i)} \bigg(   \log {\sigma(u_{w_j} \cdot v_{w_i})} + \sum_{k \ne j, k = 1}^{N_s} \log{\sigma(-u_{w_k} \cdot v_{w_i})} \bigg) \bigg].
\label{e:total_ll}
\end{align}
In practice, we use stochastic gradient descent to update the embedding vectors while minimizing the negative log-likelihood. Optimal hyper-parameters of the training algorithm including the dimensions of the embedding matrices are found using a hold-out validation set. In our model, we use $\mathcal{D} = 100$  and $N_s = 20$ based on the results hyper-parameter optimization using the hold-out set.

\subsection{Purchases vs. searches}

We fit the model described in Equation \ref{e:total_ll} separately for purchase baskets and search baskets. The reasoning from purchase baskets can be ported to browsing sessions where consumers in effect create ``search baskets'' by looking at products they intend to buy. These consideration sets are critical for our us to learn that products that are frequently bought together tend to be potential complements and products that are frequently viewed together but \textit{not} bought together are potential substitutes\footnote{Using product views to identify substitutes has also been explored in \cite{Seiler2019}}. To create consideration sets for each user session, we only include products that were viewed together but not bought together. We provide empirical evidence for relationships inferred through consideration sets later in the text.  

\subsection{Limitations}

While the method we use is novel and is able to solve the underlying problem efficiently, it is worthwhile to highlight some limitations, particularly with regard to structural models. Our method is essentially trading off ``structure'' for efficient scaling with data. Discrete choice models are guided by theory and, by focusing on a particular category, they can be used to test implications of different business strategies that the theory suggests under different assumptions. We, on the other hand, approach the problem largely from a data-driven perspective, which allows us to explore a larger product space and work with bundles from multiple categories. Further, our characterization of complementarity and substitutability (defined later) is more broad-based and we generate heuristics using machine learning that support these definitions. Again, in this case we trade-off the structure of cross-price elasticity matrices to work with a much larger set of products and learn product relationships efficiently at scale. Lastly, our model is also purposefully designed to be ``minimalistic'' that takes as input only products purchased and viewed, while ignoring all other meta information about the products. This design enables us to qualitatively validate the model and ensure that it recovers underlying consumer preferences. While we do account for other observable product characteristics when we build our supervised learning model, it is possible that adding meta information about products during the training stage can be helpful in getting better representations of products in the embedding space. We leave this task for future research.


\section{Data} \label{s:data}

We use clickstream data from a large online retailer in the US in which we observe entire user sessions of views, clicks, and purchases. The retailer sells products across multiple categories such as grocery, household, health and beauty, pet, baby products, apparel, electronics, appliances, and office supplies. The data span all consumer activity on the retailer's website from Jan-2018 to June-2018 during which we observe multiple users and multiple sessions of each user, if available.\footnote{A session is defined as a visit to the retailer's website by a user. A session continues until there is no activity by the user for 30 minutes on the website. If the user performs an action after 30 minutes of inactivity, it is considered to be a new session by the same user.} 

For each consumer's session, we observe all the products that the consumer searched or purchased along with the number of units of each product bought and the price. For all of the products, we know multiple hierarchies of the product category. The product category hierarchy can be understood using a simple example. For instance, consider the product \textit{Chobani Nonfat Greek Yogurt, Strawberry}. Its hierarchy would be Grocery ($Department$) $\rightarrow$ Dairy \& Eggs ($Aisle$) $\rightarrow$ Yogurt ($Category$), where $Department$ represents the highest hierarchy, $Aisle$ is a sub-level of $Department$, and $Category$ is a sub-level of $Aisle$ (and hence $Department$). Throughout the paper, we will refer to hierarchical categorical levels as $Department$, $Aisle$, and $Category$. In our working sample, we have products across $912$ $Categories$. It is important to note that we do not use any product meta-data for training the model. The product category hierarchy is only used for qualitatively validating the model, a point we discuss later, and generating different bundles subject to constraints on category co-membership.

As is typical of e-commerce websites, the raw clickstream data include many purchases and views of very rarely purchased products. The retailer's assortment consisted of more than 500,000 products with most products never been viewed or bought. We filter these extremely rarely purchased products to retain the top 35,000 products by product views which include more than 90\% of the purchases. After filtering, we cover about $947,000$ sessions made by $\sim 534,000$ users, which generated $\sim 861,000$ purchase baskets and $\sim 589,000$ search baskets consistings of products viewed. Observation counts from our working sample are presented in Table~\ref{t:high_level} in Appendix \ref{appendix:app_tab}. In the paper, we use the terms viewed and searched interchangeably; both imply that the user opened the description page of the product. \\

\subsection{Product baskets: Purchases and searches}

A typical user shopping session includes browsing a range of products, potentially across multiple categories, and then purchasing a subset of them. In this process, the user first forms a consideration set, i.e., a set of products from which the consumer intends to finally a choose from. In effect, from our model's perspective, the user creates two product baskets during a shopping session --- products viewed and products purchased, which form our units of analysis in this study. We distinguish between a purchased product basket and a searched product basket by including products that were purchased in the first one and viewed but \textit{not} purchased in the second one respectively. In this paper, we refer to a purchased product basket simply as purchase basket and the searched product basket as search basket. For search baskets, we only include the product in the basket if the consumer opened its detailed description page. It is important for us to distinguish between these two baskets since this separation allows us to learn different relationships between products, i.e., they could be potential complements or potential substitutes. 

\begin{figure}
	\centering
	\begin{subfigure}[t]{1.0\textwidth}
		\centering
		\includegraphics[scale=0.35]{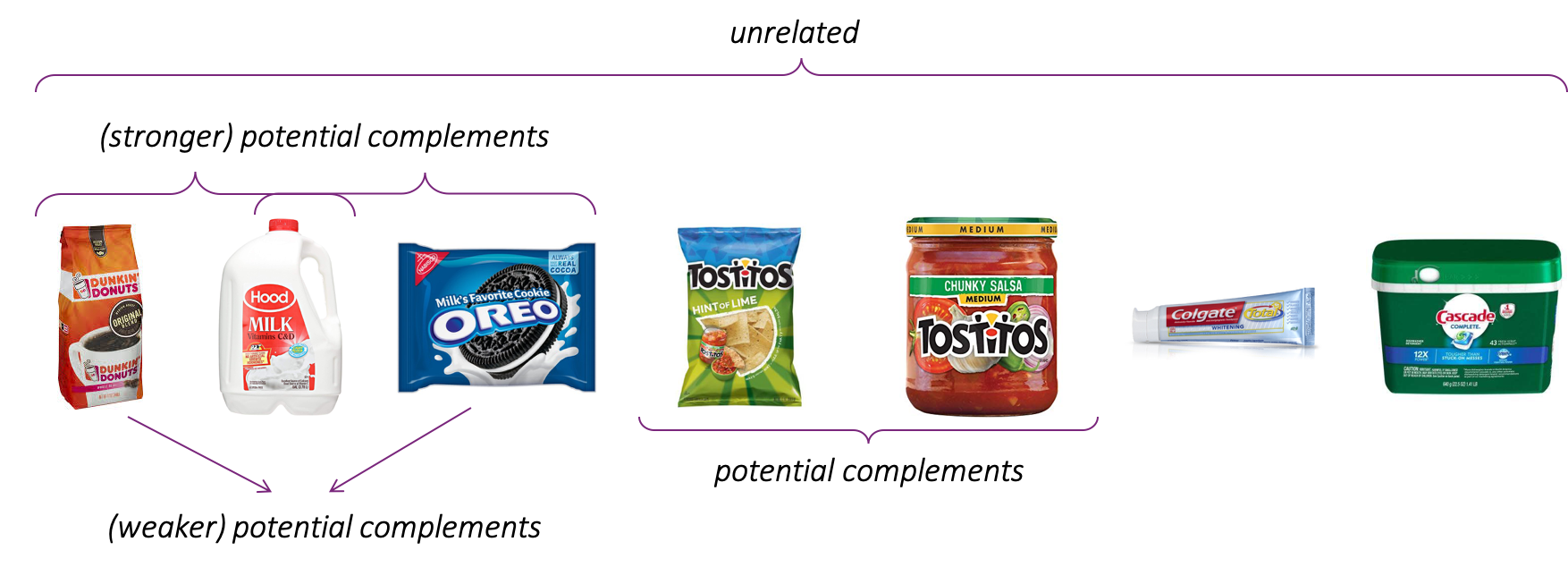} 
		\caption{Purchase basket} \label{f:pur_bask}
	\end{subfigure}
	
	\begin{subfigure}[t]{1.0\textwidth}
		\centering
		\includegraphics[scale=0.35]{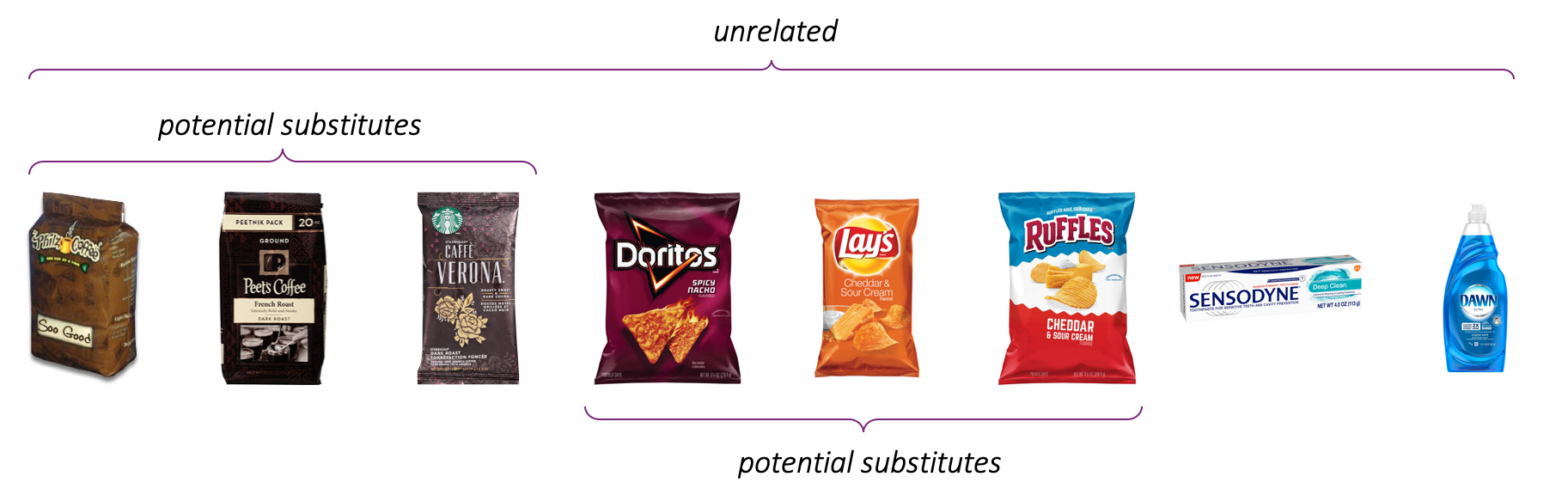} 
		\caption{Search basket} \label{f:search_bask}
	\end{subfigure}
	
	\caption{Illustrative purchase and search baskets created during a user shopping session}
	\label{f:prod_bask}
	
\end{figure}

Figure~\ref{f:pur_bask} shows an illustrative purchase basket. In this case, the user bought breakfast foods (coffee, milk, cookies), along with some snacks (chips and salsa), and two household products (toothpaste and dish pods). Our model and associated heuristics have been designed to infer that coffee, milk, and cookies are potential complements. Not only that, we want to go one step further and infer that coffee and milk are stronger complements than coffee and cookies. The signal for these relationships comes from thousands of purchase baskets where we are likely to find coffee and milk being purchased together more frequently than coffee and cookies. Similarly, we want to infer that chips and salsa are complements. The consumer in this case also purchased toothpaste and dish pods. Ex ante we do not expect any complementarity between these items and the rest of the basket and this may just be idiosyncratic noise particular to this shopping session. Note that the model does not make use of textual labels of the products. It ingests hashed product IDs and finds the relationships between these IDs, without ever looking at the product name or category.

We also observe the corresponding search basket of the same consumer, shown in Figure~\ref{f:search_bask}. The search basket includes products that were viewed but \textit{not} purchased together. We see that the consumer viewed different brands and flavors of coffee before purchasing one. Our model would infer them as potential substitutes. Further, the model would also pick out the different types of chips that the consumer searched. For inferring potential substitutes, we rely on the assumption that users search for multiple products before purchasing one, a pattern we do observe in the data.

Table \ref{t:per_visit_stats} shows the summary statistics at a basket level. On average, a consumer searches 7 products for each one bought. The mean of products bought (or viewed) is higher than the median, indicating a long right tail of baskets with many products. Within each basket, the mean number of departments is $1.6$, alluding to the concept of a focused shopping trip, i.e., a particular shopping session for groceries, a different one for household supplies, a third one for apparel, and so on. Further, we see that the average purchase basket consists of products from $3$ different categories, i.e., a consumer like Esha described in the model section could be buying groceries for breakfast from different sub-categories such as coffee, milk, and cereal. \\

\begin{table}[h]
	\centering
	\caption{Summary statistics per session}
	\scalebox{0.8}{	
		\begin{tabular}{lrrr}
			\toprule
			&      &       Purchased &  Viewed   \\
			\midrule
			\multirow{4}{2.5cm}{Products} & Mean & 3.6 & 21.3 \\
			& SD & 4.2 & 34.1 \\
			& Median &  2 &  9 \\
			& Max &  202 &  1365 \\
			\hline
			\multirow{4}{2.5cm}{Department} & Mean & 1.6 & 1.7 \\
			& SD & 0.8 & 1.0 \\
			& Median &  1 &  1 \\
			& Max &  10 &  14 \\
			\hline
			\multirow{4}{2.5cm}{Aisle} & Mean & 2.4 & 2.6 \\
			& SD & 1.9 & 2.5 \\
			& Median &  2 &  2 \\
			& Max &  28 &  48 \\
			\hline
			\multirow{4}{2.5cm}{Category} & Mean & 3.0 & 4.0 \\
			& SD & 2.9 & 5.3 \\
			& Median &  2 &  2 \\
			& Max &  69 &  122 \\
			\hline
			\multirow{4}{2.5cm}{Price} & Mean & 44.6 & 303.7 \\
			& SD & 52.1 & 526.1 \\
			& Median & 32.3 & 127.4 \\
			& Max & 2,697 & 17,480 \\
			\bottomrule
			\multicolumn{4}{p{8cm}}{\SingleSpacedXI \footnotesize{\textit{Note:} A user session is defined a visit to the retailer's website by a user. A session continues until there is no activity by the user for 30 minutes on the website. If the user performs an action after 30 minutes of inactivity, it is considered to be a new session by the same user.}} \\
		\end{tabular}
	}
	\label{t:per_visit_stats}
\end{table}


\section{Product embeddings} \label{s:embed}

We train the model described in Section \ref{s:model} using purchase and search baskets separately. Consequently, the model gives us two sets product embeddings (1) using purchase baskets that consist of product purchased together and (2) using search baskets by considering products that are viewed together but \textit{not} purchased together.\footnote{Given that our model relies solely on the co-occurrence of products within baskets, we only consider baskets that have more than $1$ product.} For training the models, we searched for optimal hyper-parameters using a hold-out sample of the data. In the paper, we report results using the models trained with the optimally tuned hyper-parameters. 

\subsection{Purchase embeddings}

\begin{figure}[h]
	\centering
	\includegraphics[scale=0.35]{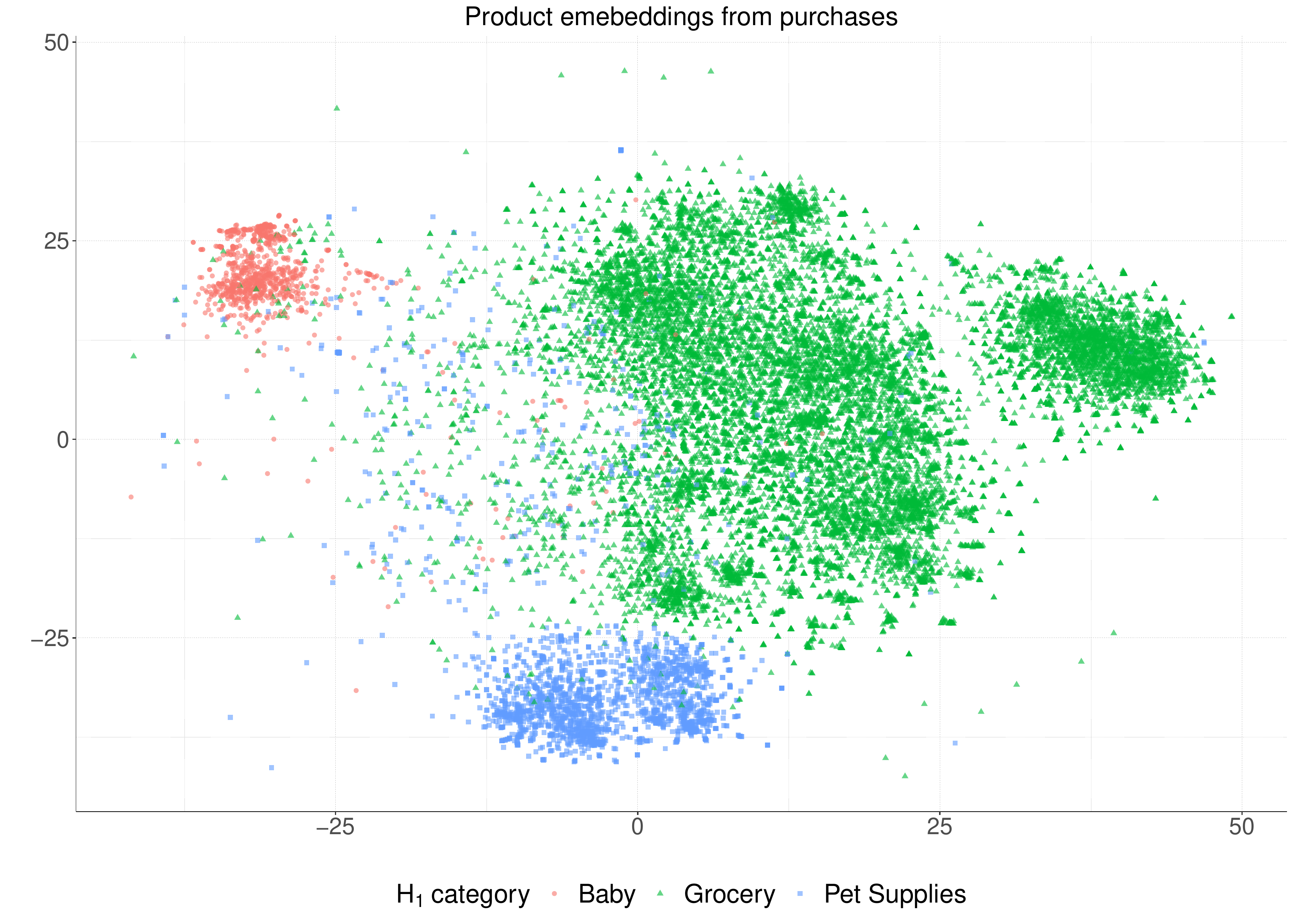}
	\caption{Product embeddings using purchase baskets}
	\label{f:pur_emb_h1}
\end{figure}

We generate purchase embeddings for products by extracting signals from their co-purchase history and condensing it to a 100-dimensional space. This is considerable gain in computational efficiency as compared to a count-based model, where a unigram model would provide a binary representation in a $35,000$-dimensional space, equal to the number of products considered. To get a perception of what the embeddings represent, we compress the embedding vectors to a 2-D space using t-SNE \citep{vanDerMaaten2008} and plot them in Figure~\ref{f:pur_emb_h1}. For visual clarity, we highlight the department of the product and show products from three departments --- groceries, baby, and pet products. A cursory glance reveals some obvious patterns. While we see a clear separation among products from different categories, there is also some overlap among departments. Again, in this space proximity to the other products indicates a higher likelihood of the two products occurring in similar contexts, or in our case, similar types of baskets. Proximity among the products in this space suggests a higher degree of complementarity\footnote{These embeddings have been plotted in a latent space and hence the scale of this axis is immaterial and only proximity between the points is important.}.

\begin{figure}[h]
	\centering
	\includegraphics[scale=0.35]{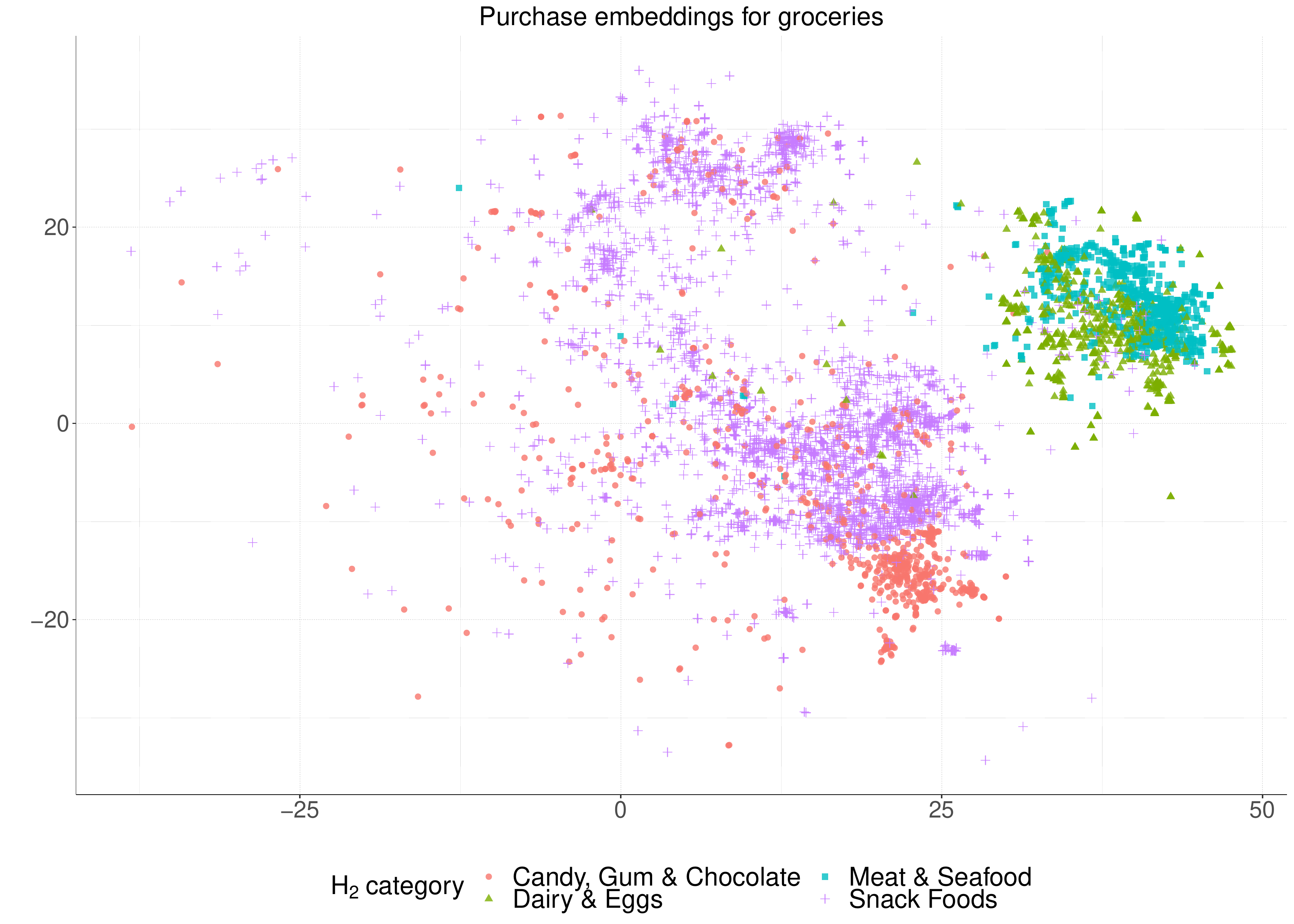}
	\caption{Purchase embeddings for products within the grocery category}
	\label{f:pur_emb_h2}
\end{figure}

As a more granular example, we zoom into the grocery department and look at snack food, meats, dairy \& eggs, and chocolates. Ex-ante we would be expect snack foods to have a stronger positive relationship with candy \& chocolates, and meats to have a stronger relationship with dairy \& eggs. Figure~\ref{f:pur_emb_h2} presents evidence for this hypothesis with snack foods being much closer to candy \& chocolates than to either meat products or dairy \& egg products. In fact, there is considerable overlap among snack foods with candy \& chocolates, suggesting a high degree of complementarity between them.

\begin{figure}[h]
	\centering
	\includegraphics[scale=0.35]{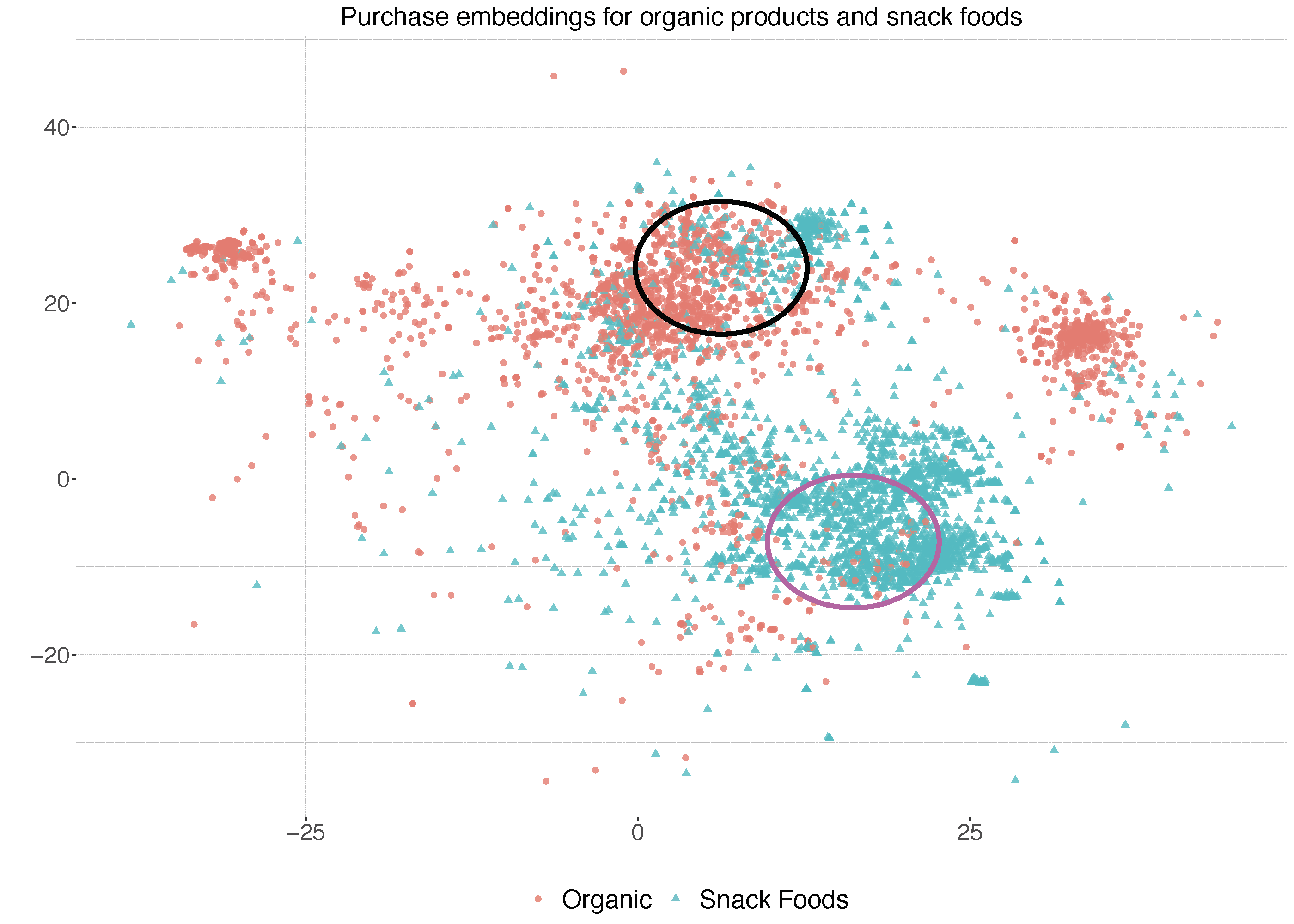}
	\caption{Purchase embeddings for organic groceries and snack foods}
	\label{f:pur_emb_org}
\end{figure}

In addition to testing relationships among products from different but pre-existing categories (typically created by the retailer), we can also generate new sub-categories of products and check how well they go with products from other categories. For instance, in Figure~\ref{f:pur_emb_org}, we compare organic groceries with snack foods. Although there is no pre-defined organic category of products, as a proof-of-concept, we do a simple string search of the word ``organic'' in the names of the products. We then visualize them along with snack foods to see what kind of organic products are related to snack foods. The upper highlighted portion of Figure~\ref{f:pur_emb_org} shows a high degree of complementarity among nuts, dried seeds such as watermelon seeds, trail mixes, jerky and dried meats, and seaweed snacks. On the other hand, the lower highlighted portion of the graph shown less of an overlap and mainly consists of cookies, chips \& pretzels. We believe that having a flexible and scalable model such as ours can provide crucial insights about market structure, brand competition, product positioning, user preferences, and personalized recommendations. We explore this idea in greater detail in a related paper. 

We dig deeper to the product level and, as examples, inspect a few focal products. Consider, for instance, organic potatoes. In the purchase space, the products closest to organic potatoes include other organic fruits and vegetables such as organic celery, organic grape tomatoes, and organic green bell peppers. Similarly, products closest to dish-washing liquid include other household items, and in some cases can be narrowed to the space of cleaning products, such as paper towels, laundry detergents, and steel cleaners. As a third example, we look at a product from the health and beauty category --- Neutrogena Oil-Free Acne Wash Redness Soothing Cream Facial Cleanser. Products that go along with this facial cleanser and include other hygiene and beauty products such as liners, rash cream, body wash, and deodorant. More details about the close complements of these focal products along with their complementarity score (described later) are presented in the Appendix~\ref{appendix:app_tab} in Tables~\ref{t:pur_sim_potato}, \ref{t:pur_sim_dish}, and \ref{t:pur_sim_face}.

We take this visual and tabular evidence as support for our claim that products that frequently co-appear in product baskets tend to have a higher degree of complementarity between them.

\subsubsection{Embeddings vs. co-purchases.}
\begin{figure}[h]
	\centering
	\includegraphics[scale=0.65]{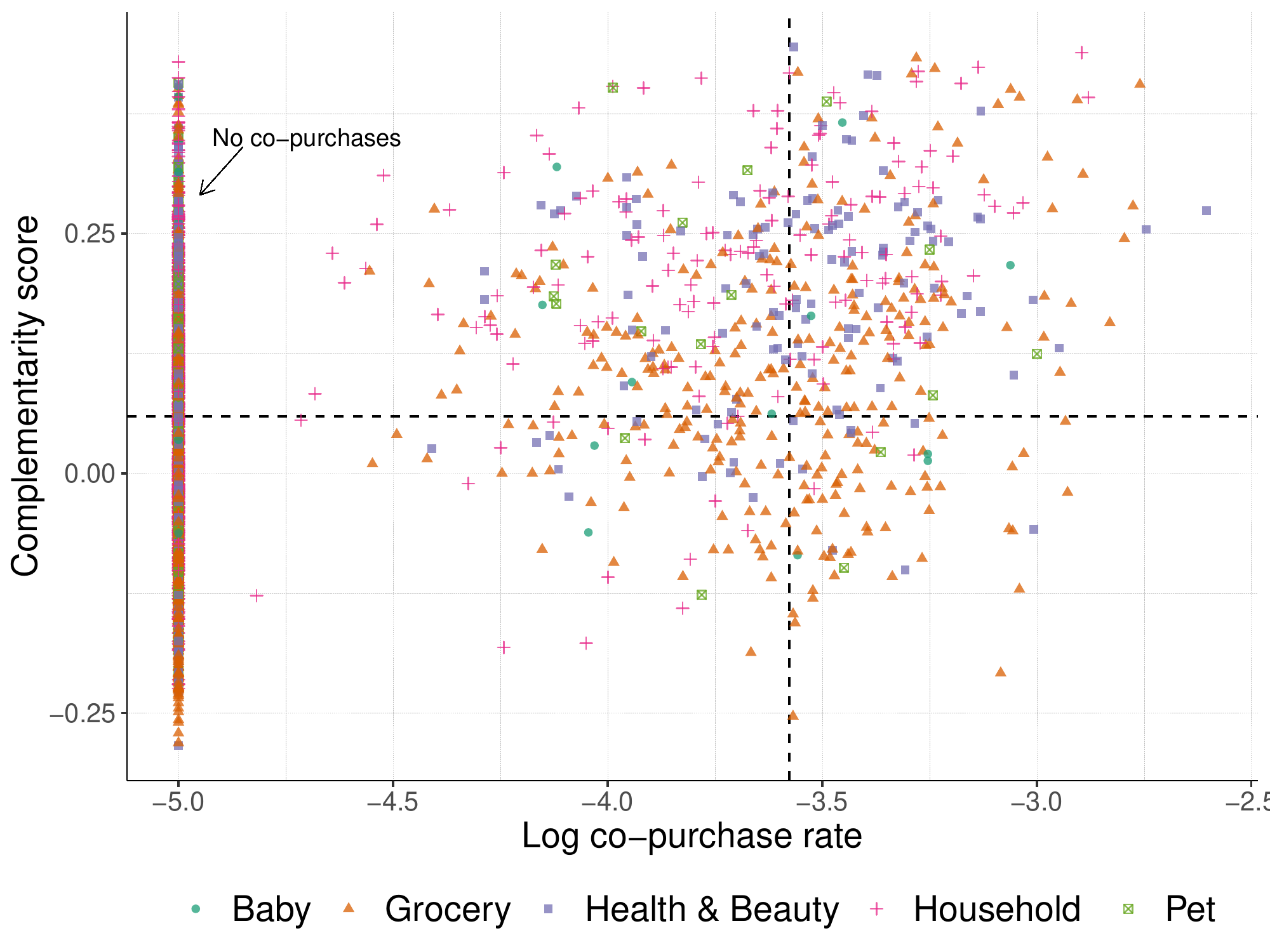}
	\caption{Historical co-purchase rate and complementarity from purchase embeddings}
	\label{f:copur_comp}
\end{figure}

A natural contender for extracting signals of complementarity is using the co-purchase frequency directly. Hence, it is worth highlighting what we obtain from the embeddings that is otherwise not available through co-purchase counts. Figure~\ref{f:copur_comp} plots (log) historical co-purchase rate as observed in the data along with a heuristic for complementarity (described later) generated from the embeddings for a random sample of 5,000 product pairs. A quick glance reveals that although co-purchases are sparse $(~10\%)$ as shown by points at the extreme left end of the graph, yet there is considerable variation in the scores of these products. This is because the model is able to smooth over the raw co-purchase counts over thousands of product pairs and learn relations even between products that have never been purchased together. This allows us to systematically explore a space that would otherwise be considered of limited value in designing bundles. Furthermore, we highlight the department of the focal product to confirm that co-purchase is a limiting factor across all top departments. 

\subsection{Search embeddings}

\begin{figure}[h]
	\centering
	\includegraphics[scale=0.35]{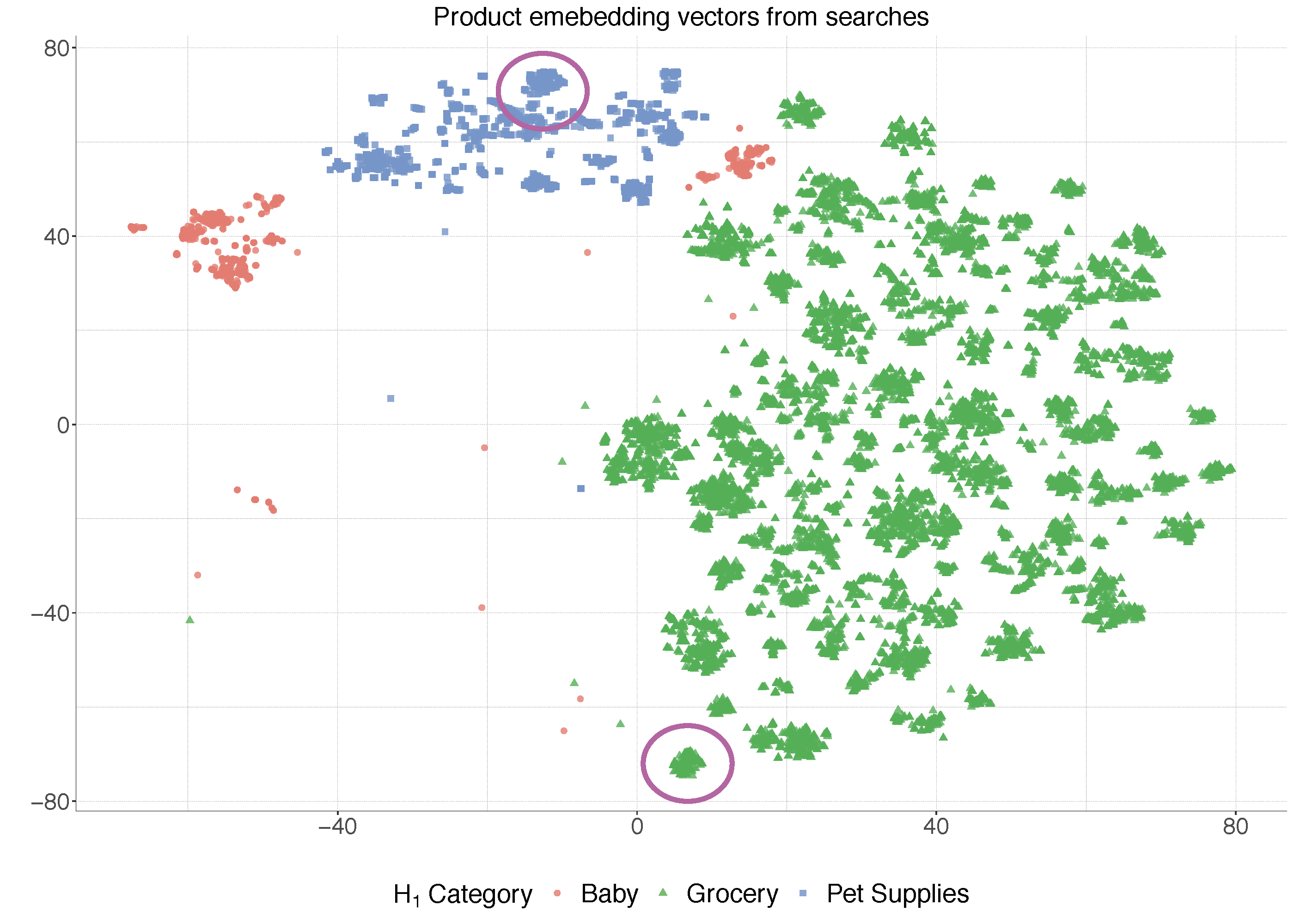}
	\caption{Product embeddings using search baskets}
	\label{f:search_emb_h1}
\end{figure}

Similar to purchase embeddings, we generate product embeddings using historical co-views of products, i.e., by observing how frequently do products co-occur in consideration set formed during thousands of shopping trips. These embeddings also lie in a similar 100-dimensional space. We condense them to a 2-D space using t-SNE \citep{vanDerMaaten2008} and plot them in Figure~\ref{f:search_emb_h1}. Again, we plot the same three departments --- groceries, baby, and pet products. The overall theme of the embeddings remains largely similar to that in purchase embeddings. However, there are two notable distinctions.  First, the inter-department clusters are further separated away indicating that most views are confined to within-department products. This reinforces the evidence we found in Table \ref{t:per_visit_stats}, where we found that most search sessions were confined to one department.  Second, there are well-defined sub-clusters within the department cluster, which self-classify into finer aisles and categories. For instance, the lowest green cluster highlighted by the purple circle comprises only of ``Breakfast Foods'' ($Aisle$), primarily containing ``Hot Cereals and Oats'' ($Category$), with occasional presence of ``Granola \& Muesli'' ($Category$). On the other end of the plot, the highlighted blue cluster on the top consists of supplies for our furry friends. This cluster only contains meat-based meals ($Category$) for dogs ($Aisle$). These observations also lend merit to our hypothesis that co-searches are good indicators of substitution across products, an insight we explore more below.

\begin{figure}[h]
	\centering
	\includegraphics[scale=0.35]{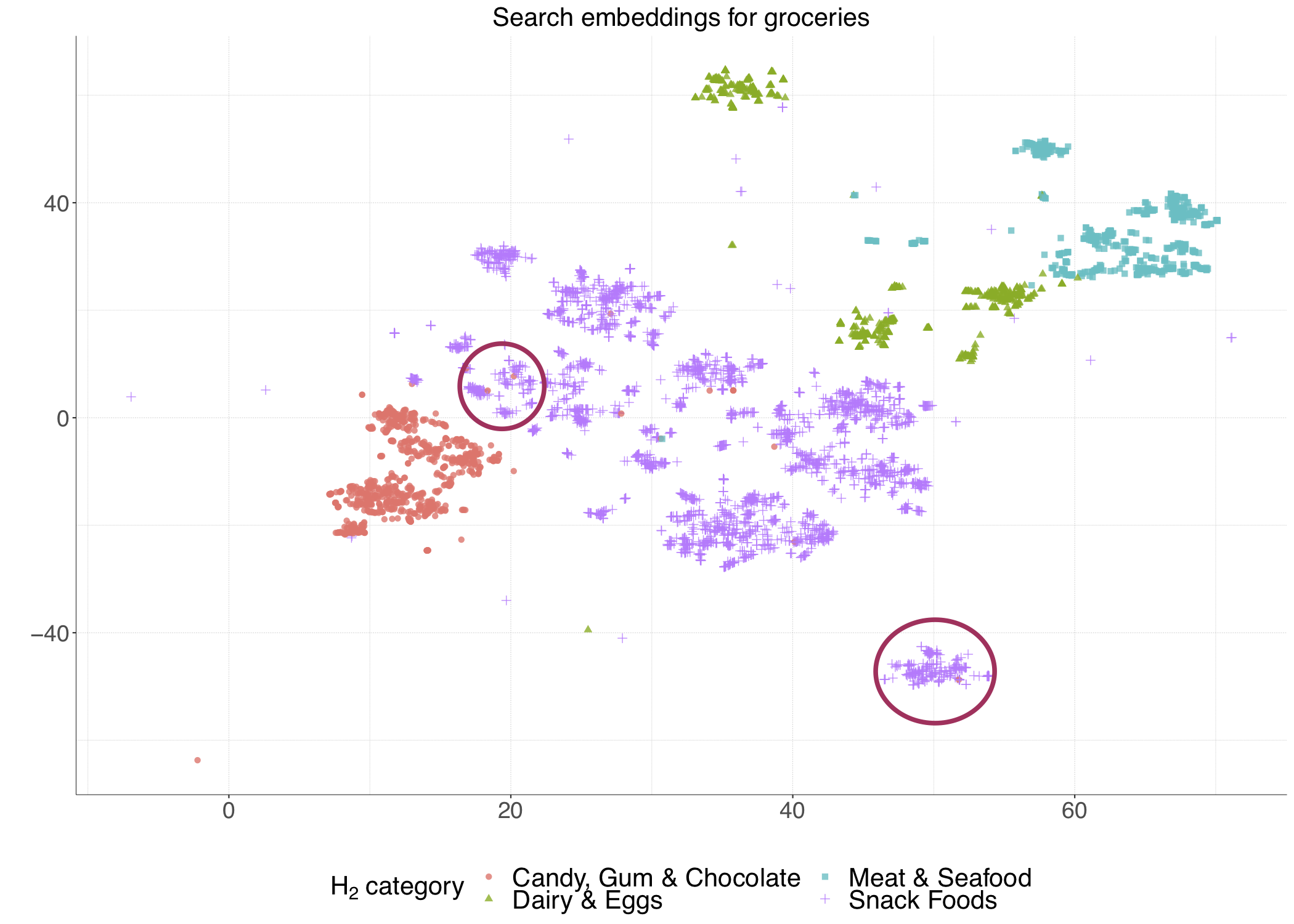}
	\caption{Search embeddings for groceries}
	\label{f:search_emb_h2}
\end{figure}

At a more granular level, we look at aisles within the grocery department in Figure~\ref{f:search_emb_h2}. We see more refined sub-clusters as compared to purchase embeddings for the same grocery products. For example, the highlighted cluster of purple points in the bottom of the graph is for popcorn and the highlighted cluster of purple points in the center left is for dried snack meats.

Similar to the purchase space, we inspect the same three products and calculate their proximity to other products in the search space. For example, organic potatoes are now closer to other varieties of potatoes in the search space such as  golden potatoes, red potatoes, and even sweet potatoes, indicating a higher degree of substitutability among them. This in contrast to the purchase space where organic potatoes were closer to other organic fruits and vegetables. Similarly, dish-washing liquid is now closer to other types and brands of dish-washing detergents such as liquids and soaps of different scents and sizes. Finally, the acne face wash shows considerable similarity with varieties of acne face washes. However, in this case there is a strong brand effect with all potential substitutes being from the same brand - Neutrogena. It could be that users have strong preferences for brands when it comes to health and beauty products or that there is a single dominant brand in product line, another point we scrutinize in greater detail in a companion paper. More examples of products closer to each other in search space are provided in the Appendix~\ref{appendix:app_tab} in Tables~\ref{t:search_sim_potato}, \ref{t:search_sim_dish}, and \ref{t:search_sim_face}.

\subsection{Product relationships}

A critical ingredient in the recipe of our bundle generation process is the relationship between any two products in the retailer's entire assortment. Furthermore, we want the relationship to be described by a metric that is continuous and category agnostic, so that we can compare the strengths of the relationships that a particular product has with other products in the assortment as well as compare strengths of the relationships across product pairs. In other words, we'd like to be able to make both within-product as well as between-product comparisons. For example, we want to be able to say that coffee and cups are stronger complements than coffee and ketchup as well as that coffee and cups are stronger complements than tea and salt. This example seems obvious, however, it becomes increasingly hard to infer these relationships when there are thousands of products in the assortments and co-purchases among pairs of products are sparse. With 35,000 products in the retailer's assortment, there are close to 50 million product combinations with over 90\% of the co-purchases being zero. Moreover, we want these relationships to be inferred from the data we observe and not be pre-imposed by the retailer. Analogously, we want to be learn that coffee and tea are stronger substitutes than coffee and fruit juice. 

With this objective in mind and based on the evidence described above, we generate a heuristic for the degree of complementarity between products $i$ and $j$ in the purchase space,
\begin{align}
	C_{ij} \triangleq  \frac{u^b_i \cdot u^b_j}{\Vert u^b \Vert \Vert u^b\Vert}, 
	\label{e:comp}
\end{align}
where $u_i^b$ and $u_j^b$ are the embeddings of products $i$ and $j$ in the purchase space respectively, and $\Vert \cdot \Vert$ is the norm of the embedding vector. This heuristic is similar to the one used by \cite{Shopper2017}. 

Similarly, we generate a heuristic for the degree of substitutability between two products $i$ and $j$ in the search space,
\begin{align}
	S_{ij} \triangleq  \frac{u^s_i \cdot u^s_j}{\Vert u^s \Vert \Vert u^s\Vert},
	\label{e:sub}
\end{align}
where $u_i^s$ and $u_j^s$ are the embeddings of products $i$ and $j$ in the search space respectively, and $\Vert \cdot \Vert$ is the norm of the embedding vector.

To give an overview of what these product relationships look like, we present examples with a few focal products. For instance, in Table \ref{t:prod_rel_ex}, we show the top-5 complements of products from three categories. In the first section, we consider hummus and we see that strong complements with it are baby carrots, greek yogurt and mandarins. On the other hand, substitutes are other varieties of hummus. Similarly, for eyeliner, we find that complementary products include other skin-care and beauty products, whereas its substitutes are other varieties of eyeliner. Lastly, for household cleaning products such as dish soap, we find other types of cleaning products as strong complements, and other varieties of dish soap as strong substitutes.\\

\begin{table}[h]
	\centering
	\caption{Product relationships using the complementarity and substitutability heuristic}
	\scalebox{0.7}{	
		\begin{tabular}{lll}
			\toprule
			Product &                               Predicted complements in the purchase space &                               Predicted substitutes in the search space\\
			\midrule
			\multirow{5}{2cm}{Sabra Classic Hummus Cups} &  Cal-Organic Baby Carrot Snack Pack &  Sabra Supremely Spicy Hummus \\
			&  Chobani Fruit On The Bottom Low-Fat Greek Yogurt &  Sabra Roasted Red Pepper Hummus Cups \\
			&  Sabra Hummus Singles &  Sabra Greek Olive Hummus \\
			&  Breakstone's 2\% Milkfat Lowfat Cottage Cheese &  Sabra Classic Hummus \\
			&  Halos Mandarins &  Sabra Hummus Singles \\
			\midrule
			\multirow{5}{2cm}{L'Oreal Paris Infallible Eyeliner} &  Chloe Eau De Parfum Spray &  L'Oreal Paris Brow Stylist Definer, Brunette \\
			&  Olay Ultra Moisture Body Wash, Shea Butter &  L'Oreal Paris Brow Stylist Definer, Dark Brunette \\
			&  John Frieda Frizz Ease Daily Nourishment Leave-In  &  L'Oreal Paris Infallible Super Slim Eyeliner, Black \\
			&  Smashbox NEW Photo Finish Foundation Primer Pore &  Milani Eye Tech Extreme Liquid Liner, Blackest Black \\
			&  Olay Quench Ultra Moisture Lotion W/ Shea Butter,  &  Revlon Colorstay Eyeliner, 203 Brown \\
			\midrule
			\multirow{5}{2cm}{Ecover Dish Soap, Pink Geranium} &   Forever New Fabric Care Wash &  Ecover Dish Soap, Lime Zest \\
			&  Ecover Fabric Softener, Sunny Day &  Earth Friendly Products Ecos Dishmate Dish Soap, Lavender \\
			&  Ecover Dish Soap, Lime Zest &  Ecover Zero Dish Soap \\
			&  Full Circle Laid Back 2.0 Dish Brush Refill &  Earth Friendly Products Ecos Dishmate Dish Soap Pear \\
			&  Giovanni Organic Sanitizing Towelettes Mixed &  Earth Friendly Products Ecos Dishmate Dish Soap Almond \\
			\bottomrule
			\multicolumn{3}{p{22cm}}{\SingleSpacedXI \footnotesize{\textit{Note:} For each focal product, the table shows the top-5 complements as determined by the embeddings in the purchase space and the top-5 substitutes as determined by the embeddings in the search space. }} \\
		\end{tabular}
	}
	\label{t:prod_rel_ex}
\end{table}

\section{Bundle generation and field experiment} \label{s:exp}

We follow a two-stage strategy to design bundles. In the first stage, we create a candidate set of bundles using the metrics of complementarity and substitutability described above and run a field experiment to gauge consumer preferences for different types of bundles. The motivation here is to develop a principled exploratory strategy that is based on a more refined action space derived using historical purchases and consideration sets. Following the field experiment, we move to the focal task in our transfer learning framework, shown in Fig~\ref{f:tl_frame}, in which we model the association between the product relationship heuristics and bundle preferences, verify the robustness of the association, and generate better bundles based on these metrics as well as certain pre-experiment co-variates. The ``learning'' happens in this stage from the mapping of metrics and the co-variates to the purchase rates of the bundles. This allows us to scale our bundle generation process to the entire assortment of products, enabling retailers to identify good candidates for bundles that consumers would prefer, thereby generating additional value for their customers.

We describe our strategy to create a candidate set of bundles for the field experiment below. In what follows, we consider bundles of two products --- a ``focal'' product and an ``add-on'' product. The focal product is the main product on whose page the bundle offer is shown and the add-on is the product on which the discount is applied. An illustrative example of how this is implemented on the retailer's website is shown in the Figure~\ref{f:pilot_example}. In order to facilitate a direct comparison between the different bundles, we offer the same relative discount on all the bundles ---  10\% off on the add-on product and full price for the focal product. The discount percentage was selected after discussions with the retailer. Before we explain the different bundle types, it is worth mentioning that the idea behind the experiment is not to horse-race different bundle types but rather learn a good strategy of making bundles as a function of the scores. By selecting bundles from different categories and departments, we intentionally add variation to explore a wider range of bundles, albeit in a principled way.

\subsection{Candidate bundles}

Our primary motivation here it to explore the potential space of bundles to generate a candidate set whose performance will be empirically validated using a field experiment. To this end, we leverage the relationships identified between products and generate a varied set of bundles. Specifically, for each focal product, we create multiple bundles across different categories, casting a wide exploratory net for learning consumer preferences, while exploiting the strength of relationships between products to guide the learning. For the field experiment, we create four types of bundles for $4,500$ products as follows:

\begin{enumerate}
	\item \textbf{Co-purchase bundles (CP):} The first category of bundles is based on high observed co-purchase frequency. For each focal product, we select the product that it has been most frequently co-purchased with. These bundles are the natural contenders for a simple data-driven bundling strategy - products that have been purchased frequently together in the past will have a higher likelihood of being purchased together in the future as well, \textit{ceteris paribus}. They also serve as a useful starting point of our bundle design strategy since we can map these bundles back to the underlying complementarity and substitutability scores, allowing us to learn more generalized patterns. However, these bundles are limited in scope since this strategy (a) does not generate bundles of products that have never been co-purchased before, (b) uses co-purchase information even when it is very noisy, e.g., bundling products if they have been co-purchased, say, 2 times in the past, (c) does not explore cross-category options since most of the bundles come from the same categories and aisles. An example of this type of bundle is toothpaste and toothbrush.
	
	\item \textbf{Cross-category complements (CC):} For a focal product $i$, we consider the strongest complement for $i$ across a different category but within the same department. The idea behind this strategy is to identify products that are most likely to be complements in usage and hence having the focal product under consideration would indicate a high-likelihood of purchasing the add-on product as well. However, to add an element of exploration, we pair products across different categories. In case of a tie with the above co-purchase bundles, we use the second strongest complement. A simple example of this is bundling toothpaste and mouthwash.
	
	\item \textbf{Cross-department complements (DC):} These bundles are similar in spirit to the cross-category complementary bundles mentioned above except that they specifically search over departments that are different from that of the focal product. Since most purchases within a trip come from the same department, as shown in Table~\ref{t:per_visit_stats}, we tend to find stronger complements within the same department. Hence, the motivation in this arm is to explore cross-department bundles (e.g., household supplies and beauty products) of products that would otherwise not be considered. An example for this would be bundling toothpaste and night cream together.
	
	\item \textbf{Variety (VR):} Extant research has suggested the benefit of bundling (imperfect) substitutes to capture a larger portion of the consumer surplus and improve profitability \citep{Lewbel1985, Venkatesh2003}. We explore this idea empirically by creating bundles of products that are close to each other in the search space. The rationale here is that products that appear to be potential substitutes may in fact also be complements over time or complements within a household. If this is true, then bundling products that are imperfect substitutes could help exploit variety seeking behavior among consumers and generate incremental sales for the retailer. For example, bundling two different varieties or flavors of toothpaste.\footnote{There is obviously the caveat that consumers may actually be forward looking and just buy the products ahead of time while they are being sold at a discount thereby having no impact on overall sales of the retailer. We do not investigate inter-temporal substitution patterns in this study while noting that it is a interesting avenue to study further.}
\end{enumerate}

We create $18,000$ bundles across the different types mentioned above. Basic characteristics of the bundles across the four types are shown in Table~\ref{t:bun_cat}. We show the results for $9,728$ bundles which were viewed at least once during the experiment, and hence are part of our subsequent analysis. All values in this table are calculated based on the pre-experiment data used for training. The number of bundles viewed is different across the four bundle types since there is flux in the inventory and depending upon the location and time of the consumer, a bundle may or may not be available. We also provide a similar table using focal products that have all the different bundle types viewed in Table~\ref{t:bun_cat_complete} in Appendix~\ref{appendix:app_tab}.

We calculate the co-purchase rate for the product pairs. \textit{Price-1}, \textit{Rating-1}, \textit{Purchase rate-1} correspond to the average price of the focal products, their average user provided ratings, and their historical individual purchase rates. Analogously, \textit{Price-2}, \textit{Rating-2}, and \textit{Purchase rate-2} correspond to the same variables for the add-on product. The last three rows show the mean of binary variables which take the value 1 if both the product belong to the same brand, same aisle, and the same category respectively. 

\begin{table}[h] \centering 
	\caption{Bundle types based on relationship heuristics} 
	\label{t:bun_cat} 
	\scalebox{0.9}{	
		\begin{tabular}{@{\extracolsep{5pt}} lrrrr} 
			\\[-1.8ex]\hline 
			\hline \\[-1.8ex]
			&  Co-purchase &  Cross  &   Cross  & Variety \\ 	
			&   &  category &   department  &  \\ 	
			\cmidrule{2-5} \\
			&  (CP) &  (CC) &    (DC) & (VR) \\ 	
			\hline \\[-1.8ex] 
			Bundles viewed & $2,189$ & $2,632$ & $2,126$ & $2,781$ \\ 
			Comp. score & $0.40$ & $0.34$ & $0.22$ & $0.40$ \\ 
			Sub. score & $0.59$ & $0.38$ & $0.19$ & $0.74$ \\ 
			Co-purchase rate & $0.58$ & $0.22$ & $0.06$ & $0.42$ \\ 
			Price-1 & $9.03$ & $9.57$ & $10.22$ & $9.15$ \\ 
			Price-2 & $8.23$ & $7.70$ & $10.80$ & $9.27$ \\ 
			Purchase rate-1 & $0.04$ & $0.04$ & $0.04$ & $0.04$ \\ 
			Purchase rate-2 & $0.04$ & $0.05$ & $0.06$ & $0.04$ \\ 
			Product rating-1 & $4.70$ & $4.71$ & $4.73$ & $4.70$ \\ 
			Product rating-2 & $4.73$ & $4.71$ & $4.81$ & $4.70$ \\ 
			Same brand & $0.45$ & $0.16$ & $0.10$ & $0.47$ \\ 
			Same aisle & $0.72$ & $0.56$ & $0$ & $0.99$ \\ 
			Same category & $0.53$ & $0.01$ & $0$ & $0.97$ \\ 
			\hline \\[-1.8ex] 
			\multicolumn{5}{p{11.5cm}}{\SingleSpacedXI \footnotesize{\textit{Note 1:} Co-purchase rate has been multiplied by 100. Price-1, Purchase rate-1, and Product rating-1 show the average price, average historical purchase rate, and the average product rating of the focal product in each bundle type. Price-2, Purchase rate-2, and Product rating-2 are corresponding variables for the add-on product. Same brand, Same aisle, and Same category are binary variables that indicate if the two products are from the same brand, same aisle, and the same category respectively.}} \\
			\multicolumn{5}{p{11.5cm}}{\SingleSpacedXI \footnotesize{\textit{Note 2:} For 26 bundles in the CC type, the add-on product category had been incorrectly recorded in the retailer's database. We note their correct category here and all subsequent analysis is done with the correct category.}} \\
		\end{tabular} 
	}
\end{table}

Finally, we note that our focus here is on identifying the best promotional bundles for consumers and we do not explicitly optimize for profitability or revenue maximization, which in itself is a challenging computational pursuit. However, we do put reasonable constraints on the bundles we create after deliberations with the retailer. Specifically, we only create bundles containing products with net positive margin after including the 10\% discount.

\subsection{Field experiment}
	
Our algorithm to create product embeddings and learn product relationships allows us to generate a wide set of candidate bundles that consumers would prefer. Our goals with the field experiment are to empirically validate how different bundles perform and learn high-level strategies that can be effectively implemented by managers. 

We ran the field experiment for two months from mid-July 2018 to mid-September 2018. The experiment was run at a user-product level, such that if a user $m_1$ searched for product $i$ which has a bundle associated with it, then the user would be randomized into one of the four treatments, i.e., the consumer would be shown one of the four bundles associated with the focal product. Let's say that user $m_1$ was randomized into the cross-category complement bundle arm for product $i$, then every time $m_1$ searched for $i$, she would be offered the opportunity to buy the cross-category complement bundle with the discount. The user need not buy the bundle and can still purchase either the focal product directly or the add-on product without any discount. After searching for $i$, if $m_1$ searched for product $j$, she would again be randomized into any of the four treatments. However, if she searched for $i$ again, she would be in the same cross-category complement treatment. To give a perspective of how the bundle offer is presented to the user, we show two illustrative examples in Figure~\ref{f:pilot_example}.

As expected due to the randomization, pre-experiment covariates --- number of visits, number of product views, number of different products added-to-cart, total number of units purchased (accounting for multiple units of the same product purchased), and total revenue --- are statistically indistinguishable across the different bundle types (Appendix, Table~\ref{t:pre_exp_norm}).
	
An overview of the results from the field experiment is shown in Table \ref{t:pilot_kpi}. $9,728$ bundles were viewed a total of $356,368$ times by $164,469$ users during the experiment. A visit to a bundle is the same as the visit to the focal product (as shown in Figure~\ref{f:pilot_example}). We also capture clicks on the bundle component on the web page, the number of bundles added-to-cart (ATC), and bundle purchases. The third column shows the same metrics as a proportion of the total number of bundle views.

\begin{table}[h] \centering 
	\caption{Key metrics from the field experiment} 
	\label{t:pilot_kpi} 
	\scalebox{0.9}{	
		\begin{tabular}{@{\extracolsep{5pt}} lrr} 
			\\[-1.8ex]\hline 
			\hline \\[-1.8ex] 
			& Count & Count/ \\ 
			& 			&  Views \\ 
			\hline \\[-1.8ex] 
			Unique bundles viewed & $9,728$ & -  \\ 
			Total bundle views & $356,368$ & - \\ 
			Bundle clicks & $5,197$ & $0.015$ \\ 
			Bundle ATC & $2,847$ & $0.008$ \\ 
			Bundle purchases & $503$ & $0.001$ \\
			\hline \\[-1.8ex] 
			\multicolumn{3}{p{7cm}}{\SingleSpacedXI \footnotesize{Note: The third column is the second column divided by the total number of views. ATC is add-to-cart.}} \\
		\end{tabular} 
	}
\end{table} 

We further investigate the results split by bundle type. Table~\ref{t:bp_metrics} shows variation in the views, clicks, and purchases of bundles across the different bundle types. We focus on two metrics of success for the bundles --- the add-to-cart rate and the purchase rate. Add-to-cart (ATC) rate is the ratio of add-to-cart events and total views and the purchase rate is the ratio of bundle purchases to total views. These rates are largely highly statistically significantly different between pairs of types (Table~\ref{t:prop_test} in the Appendix). 
Similar results using only focal products that have a bundle observed for each of the four types are provided in the Appendix in Table~\ref{t:bp_metrics_complete}. 
	
To reiterate, the aim here was to try a refined sampling of bundles so as to explore the action space and learn bundle success likelihood as a function of the underlying scores. Consequently, we don't dive too deep into the comparative results of the experimental bundle types. However, we do note a few interesting points. First, as is expected, the co-purchase bundles (CP), consisting of products that have been frequently purchased together in the past, tend to do quite well. Adding a discount to frequently co-purchased bundles would have further increased their likelihood of purchase. However, as mentioned earlier, most of these bundles come from the same aisle and do not exploit the range of the retailer's assortment. The embeddings allows us to tap into that variation systematically by mapping these bundles to the underlying product relationship scores. We describe this process in greater detail in the next section. 

Variety bundles (VR) have a high purchase rate as well and their performance is similar to co-purchase bundles (CP). The cross-category complements (CC) have a lower purchase rate as compared to co-purchase and variety bundles. Further, while the cross-department (DC) bundles did not perform as well as the other categories but they provide interesting insights into cross-department promotion strategies. For example, a popular cross-department bundle was between household supplies and skin care --- laundry detergent + hand soap. Other popular cross-department bundles include coffee + paper napkins, and  shower cleaner + whitening toothpaste. Correspondingly, popular cross-category bundles were granola bars + crackers, mops + floor cleaners, and disposable razors + body wash. Popular variety bundles mostly consisted of close, but still imperfect, substitutes such as frozen meals with vegetable korma + Bombay potatoes,  snacks such as organic pumpkin seeds + organic raw almonds, and pasta with gluten free rotini + gluten free penne. In fact, grocery variety bundles performed particularly well. A few more examples of the best performing bundles from each type are shown in Table~\ref{t:top_3_btype}.
	
\begin{table}[h] \centering 
	\caption{Experiment results split by bundle type} 
	\label{t:bp_metrics} 
	\scalebox{0.9}{
		\begin{tabular}{@{\extracolsep{5pt}} lrrrr} 
			\\[-1.8ex]\hline 
			\hline \\[-1.8ex] 
			  &  Co-purchase &  Cross-cat.  &   Cross-dept.  & Variety \\ 	
			    \cmidrule{2-5} \\
			    &    (CP)           &  (CC)     & (DC)  & (VR) \\ 
			\hline \\[-1.8ex] 
			Bundles & $2,189$ & $2,632$ & $2,126$ & $2,781$ \\ 
			Views & $94,458$ & $88,757$ & $81,239$ & $91,914$ \\ 
			Clicks & $1,586$ & $1,014$ & $794$ & $1,803$ \\ 
			ATC & $1,050$ & $665$ & $289$ & $843$ \\ 
			Purchases & $198$ & $102$ & $47$ & $156$ \\ 
			\hdashline \\
			CTR & $0.017$ & $0.011$ & $0.010$ & $0.020$ \\ 
			ATC rate & $0.011$ & $0.008$ & $0.004$ & $0.009$ \\ 
			Purchase rate & $0.002$ & $0.001$ & $0.001$ & $0.002$ \\ 
			\hline \\[-1.8ex] 
			\multicolumn{5}{p{8.5cm}}{\SingleSpacedXI \footnotesize{\textit{Note 1:} CTR is click-through rate. ATC is add-to-cart. The rate columns in the right half of the table are calculated as a proportion of views.}} \\
		\end{tabular}
		}
\end{table} 

The field experiment serves an intermediate step that bring us closer to the focal task in our transfer learning framework. By systematically adding variation across different bundle types we are able to learn consumer preferences across a range of bundles coming from multiple categories of the retailer's assortment. However, by themselves, individual bundle results are quite noisy and do not directly lend themselves towards implementable insights. In the next section we tie the relationship scores from the embeddings with the results from the experiment using predictive modeling to derive more general and robust findings.

\section{Generalization with supervised learning} \label{s:predict}

\begin{figure}[h]
	\centering
	\begin{subfigure}{.5\textwidth}
		\centering
		\includegraphics[scale=0.4]{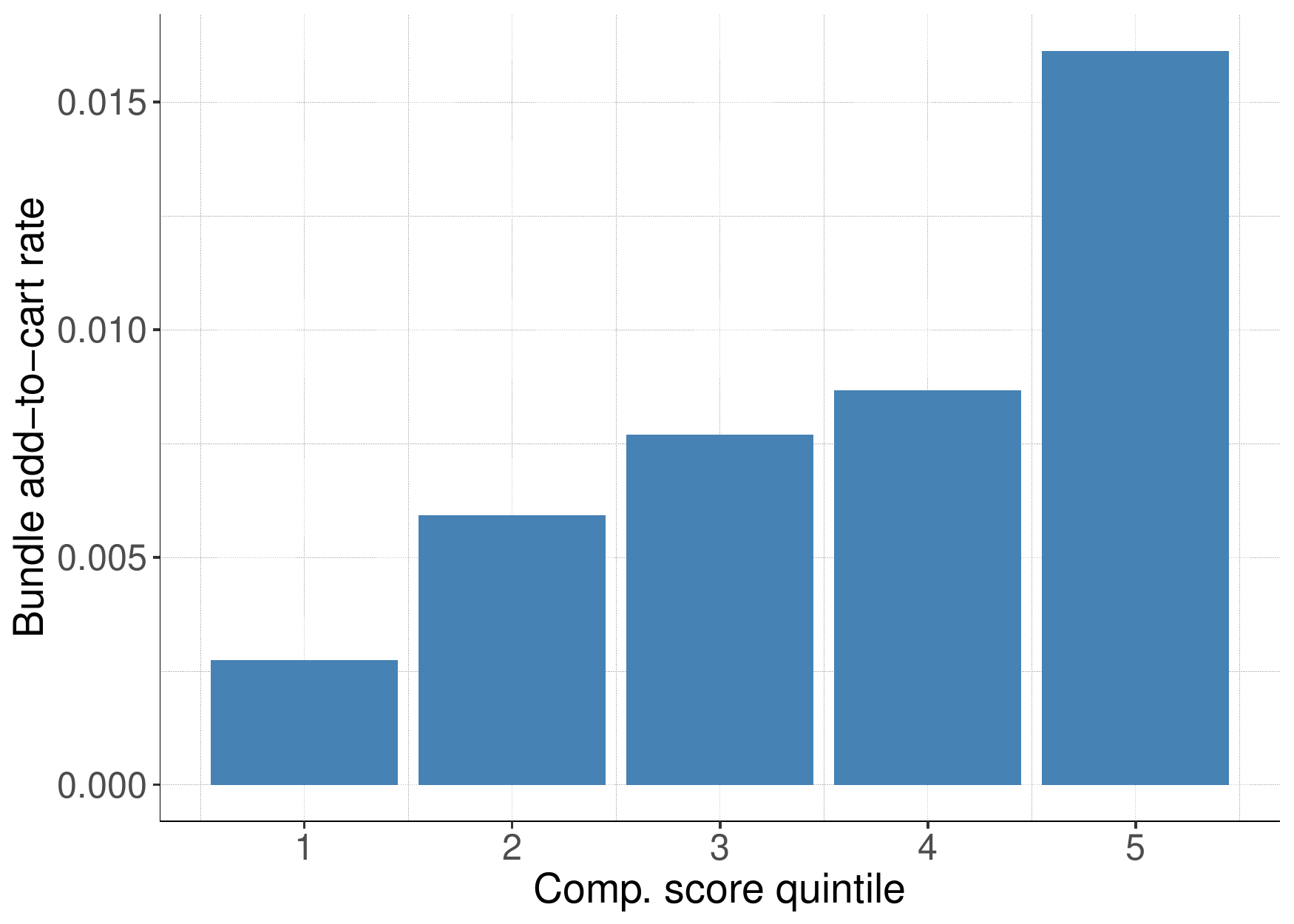}
		\caption{Complementarity score}
		\label{f:comp_atc_bar}
	\end{subfigure}%
	\begin{subfigure}{.5\textwidth}
		\centering
		\includegraphics[scale=0.4]{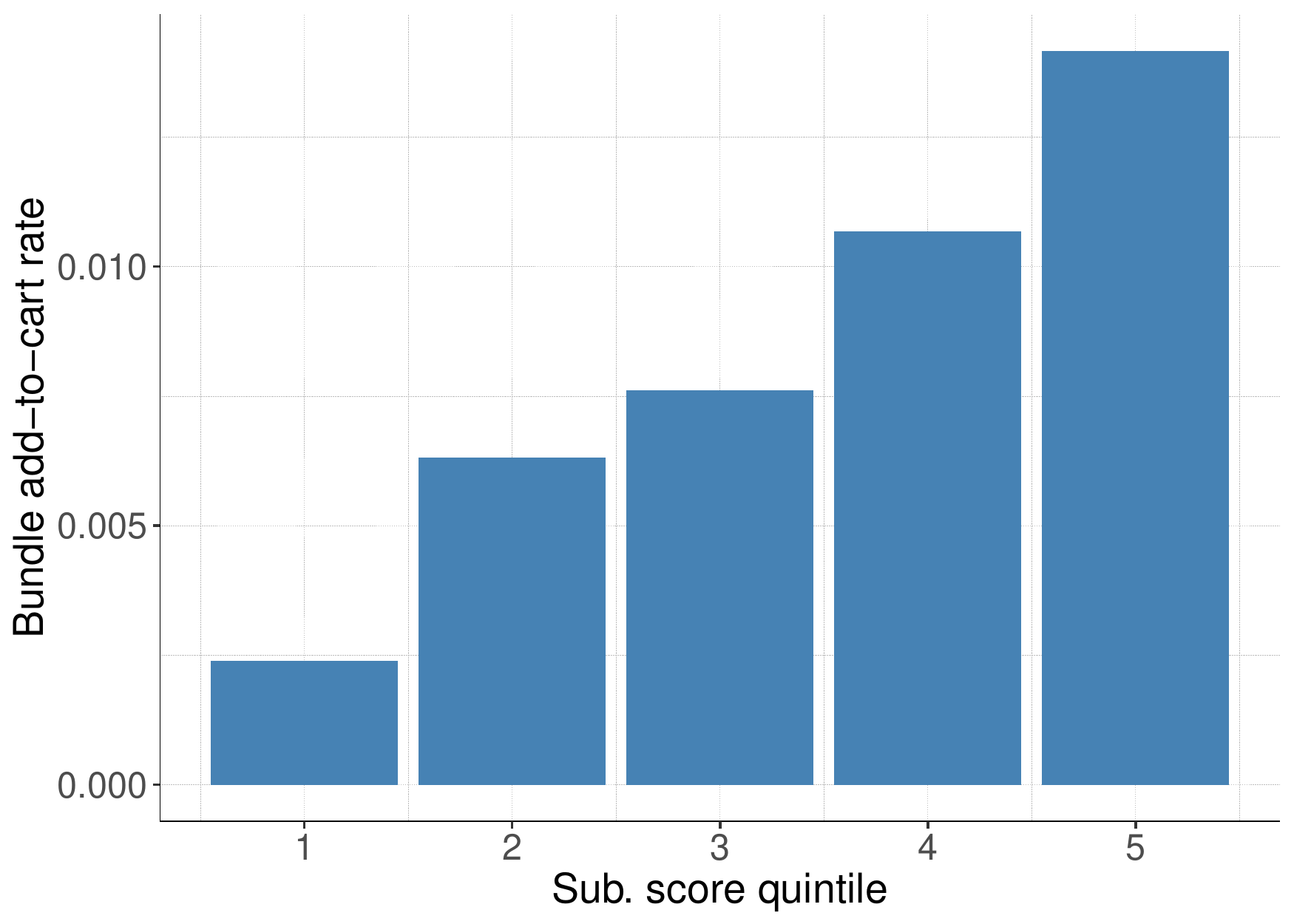}
		\caption{Substitutability score}
		\label{f:sub_atc_bar}
	\end{subfigure}
	\caption{Bundle add-to-cart rate as a function of relationship scores}
	\label{f:total_atc_bar}
\end{figure}

Results from the field experiment show that there is value in using the underlying product embeddings to generate product bundles. This is coarsely illustraed in Figure~\ref{f:total_atc_bar}, which shows the performance of the bundles across quintiles of the two relationship heuristics. The distribution of the scores split by product departments is shown in the Appendix in Figure~\ref{f:box_l0_atc}. While this suggests that the heuristics are a good approach to form promotional bundles, it also is a fertile exploratory ground to identify more, and perhaps even better, bundles. This brings us to the focal task in our transfer learning framework --- developing a supervised learning model to harness the value in the product relationships and predict bundle success likelihood. 

Mathematically, a general version our focal task is shown in Equation~\ref{e:tf_1}, where $Y_{ij}$ is the outcome observed (bundle added-to-cart or not) in the field experiment for a bundle consisting of the focal product $i$ and the add-on product $j$:
\begin{align}
Y_{ij} = f \big(\hat{u}^b_i, \hat{u}^b_j, \hat{u}^s_i, \hat{u}^s_j, W_{ij} \big) + \epsilon_{ij},
\label{e:tf_1}
\end{align}
where
$\hat{u}^b_i, \hat{u}^b_j, \hat{u}^s_i, \; \text{and} \; \hat{u}^s_j$ are the embeddings of the products estimated using the model described in Section~\ref{s:model}. We essentially ``transfer'' the knowledge of product relationships learned through the embeddings to predict bundle add-to-cart, a different albeit related task. $W_{ij}$ includes the pre-treatment variables and product attributes. The task is two-fold then: choose $f$ and estimate its parameters. The choice of $f$ depends on high-level objective of the learning exercise. If the goal is solely prediction then $f$ can take a broad range of forms including a neural network, which, in addition, would also allow us to update the embeddings through backpropagation. This will further fine-tune the embeddings and improve their predictive power towards the focal task. In our case, we are more interested in inference and verifying the robustness of the product relationships in predicting bundle success. Hence, we choose a more interpretable approach and build a hierarchical logistic regression model, allowing the intercept and slopes for the product relationship variables to vary at the aisle level, as shown in Equation~\ref{e:hglm}.
\begin{align}
Pr(Y_{ij} = 1 \vert \text{View}_i) &= \text{logit}^{-1}\bigg(\alpha_{k[ij]} + \beta^{b}_{k[ij]} C_{ij} + \beta^{s}_{k[ij]} S_{ij} + \gamma^T W_{ij} \bigg) \\
\begin{pmatrix}
\alpha_k    \\
\beta^b_k  \\
\beta^s_k \\
\end{pmatrix} & \sim N\left(\begin{pmatrix}
\mu_{\alpha} \\
\mu_{\beta^b} \\
\mu_{\beta^s} \\
\end{pmatrix}, \Sigma \right)
\label{e:hglm}
\end{align} 
where $i$ indexes the focal product, $j$ is the add-on product, and $k$ is the aisle of the focal product and hence, depends on $i$. Together, $ij$ make one bundle and we estimate the probability of the bundle being added-to-cart, conditional on the user viewing the focal product $i$. The intercept $\alpha_k$ is allowed to vary by aisle along with the slopes $\beta^b$ and $\beta^s$, i.e., the coefficients for the complementarity heuristic $C_{ij}$ and the substitutability heuristic $S_{ij}$. Pre-treatment variables and other product meta-data are captured by the vector $W$ with their parameters $\gamma$ held fixed. The varying parameters are estimated jointly with each parameter having a separate mean and variance. Their variances and covariances are given by the $3 \times 3$ matrix $\Sigma$. In what follows, we use the terms ``aisle'' and ``category'' interchangeably for generalizable insights although the statistical model and analysis is done at the aisle level.  

Setting up the model in a hierarchical fashion helps us account for unobserved variation in product aisles that is not captured in a pooled regression model. Further, since we model the product aisles separately, we can use the model to generate predictions for aisles that were not part of our training data, and hence generalize our findings to a larger domain. In addition to statistical benefit, the hierarchical modeling also provides a systematic way to analyze the robustness of our model across aisles, as we show later. 	

To re-emphasize, the statistical objective here is to map the product relationship scores and other pre-treatment variables to a bundle performance measure such as bundle add-to-cart. However, the more substantive objectives behind setting up this framework is to(1) investigate the predictive power of the heuristics for bundle success after controlling for pre-experiment variables and product meta-data, (2) verify their robustness against different specifications, (3) derive generalizable insights to help better understand the bundle design process, and (4) efficiently create more effective bundles using other products in the retailer's assortment. We do this exercise by running a hierarchical logistic regression to predict a user's likelihood to add a bundle to their cart, conditional on the user viewing the bundle. The model is run at a bundle level by defining the successes as the total number of unique users who added the bundle to their cart and the failures as the total number of users who viewed the bundle but did not add-to-cart. 

The results of the regression are shown in column 4 of Table \ref{t:reg_both_re}. For comparison, we show three other models: (1) using only the scores, (2) using the scores and pre-treatment variables while controlling for Aisle fixed effects, and (3) a hierarchical model with the intercept, but not other coefficients, varying at the focal product aisle level. In models 2, 3, and 4, we control for the historical co-purchase rate, prices of both products, signed and squared difference between the two prices, the rating of both products, their individual historical purchase rates. Further, we include dummy variables to account for whether the products are from the same brand, belong to the same category, and whether they belong to a different category conditional on being from the same aisle. Standard errors are clustered at the focal product level. Summary statistics for the variables in the model are shown in the Appendix in Table~\ref{t:sum_stat_model}.

\begin{table}[h] \centering 
	\caption{Hierarchical logistic regression to predict bundle add-to-cart} 
	\label{t:reg_both_re} 
	\scalebox{0.8}{ 
		\begin{tabular}{@{\extracolsep{5pt}}lcccc} 
			\\[-1.8ex]\hline 
			\hline \\[-1.8ex] 
			& \multicolumn{4}{c}{\textit{Dependent variable:}} \\ 
			\cline{2-5} 
			\\[-1.8ex] & \multicolumn{4}{c}{Bundle add-to-cart} \\ 
			\\[-1.8ex] & \multicolumn{2}{c}{\textit{Logistic}} & \multicolumn{2}{c}{\textit{Hierarchical logistic}} \\ 
			& \multicolumn{2}{c}{\textit{}} & \multicolumn{2}{c}{\textit{regression}} \\ 
			\\[-1.8ex] & (1) & (2) & (3) & (4)\\ 
			\hline \\[-1.8ex] 
			Comp. score & 0.230$^{***}$ & 0.251$^{***}$ & 0.244$^{***}$ & 0.308$^{***}$ \\ 
			& (0.039) & (0.041) & (0.039) & (0.062) \\ 
			Sub. score & 0.246$^{***}$ & 0.153$^{**}$ & 0.137$^{**}$ & 0.171$^{**}$ \\ 
			& (0.044) & (0.072) & (0.064) & (0.067) \\ 
			Hist. co-purchase rate &  & 0.048$^{*}$ & 0.048$^{***}$ & 0.066$^{***}$ \\ 
			&  & (0.028) & (0.018) & (0.021) \\ 
			Price-1 &  & $-$0.012$^{**}$ & $-$0.014$^{***}$ & $-$0.015$^{***}$ \\ 
			&  & (0.005) & (0.004) & (0.004) \\ 
			Price-2 &  & $-$0.025$^{***}$ & $-$0.025$^{***}$ & $-$0.022$^{***}$ \\ 
			&  & (0.005) & (0.005) & (0.005) \\ 
			Rel. price interaction &  & 0.144$^{***}$ & 0.155$^{***}$ & 0.167$^{***}$ \\ 
			&  & (0.049) & (0.058) & (0.055) \\ 
			Hist. Purchase Rate-1 &  & $-$1.155 & $-$1.224 & $-$0.905 \\ 
			&  & (1.441) & (1.163) & (1.198) \\ 
			Hist. purchase rate-2 &  & 2.066$^{*}$ & 2.035$^{**}$ & 2.216$^{**}$ \\ 
			&  & (1.166) & (0.863) & (0.871) \\ 
			Rating-1 &  & 0.088 & 0.111 & 0.100 \\ 
			&  & (0.093) & (0.082) & (0.083) \\ 
			Rating-2 &  & $-$0.033 & $-$0.031 & $-$0.037 \\ 
			&  & (0.083) & (0.072) & (0.072) \\ 
			Same brand &  & 0.310$^{***}$ & 0.335$^{***}$ & 0.356$^{***}$ \\ 
			&  & (0.091) & (0.072) & (0.074) \\ 
			Same category &  & 0.281$^{**}$ & 0.314$^{***}$ & 0.286$^{**}$ \\ 
			&  & (0.132) & (0.113) & (0.115) \\ 
			Diff. category \& same aisle &  & 0.417$^{***}$ & 0.440$^{***}$ & 0.437$^{***}$ \\ 
			&  & (0.106) & (0.091) & (0.092) \\ 
			Constant & $-$4.931$^{***}$ & $-$4.493$^{***}$ & $-$5.452$^{***}$ & $-$5.429$^{***}$ \\ 
			& (0.039) & (0.666) & (0.501) & (0.501) \\ 
			\hline \\[-1.8ex] 
			Aisle-specific intercepts &  & \checkmark & \checkmark & \checkmark \\ \\[-1.8ex] 
			Aisle-specific slopes &  &  &  & \checkmark \\ 
			\hline \\[-1.8ex] 
			Observations & 9,728 & 9,728 & 9,728 & 9,728 \\ 
			\hline 
			\hline \\[-1.8ex] 
			\textit{Note:}  & \multicolumn{4}{r}{$^{*}$p$<$0.1; $^{**}$p$<$0.05; $^{***}$p$<$0.01} \\ 
			\hline 
			\hline \\[-1.8ex] 
			\multicolumn{5}{p{12.5cm}}{\SingleSpacedXI \footnotesize{\textit{Note:} The models predict the likelihood of bundle purchase conditional on the bundle being viewed. The first column uses both the scores between the two products. The second model controls for historical co-purchase rate, prices, individual historical purchase rates of both the products, product ratings, a dummy if the products are from the same brand, a dummy if the products belong to the same product category, and a dummy if the products belong to the same aisle but different category. The third column allows the intercept to vary at the level the aisle of the focal product, and the fourth column allows the intercept and the slopes for the scores to vary at the level the aisle of the focal product. Continuous variables have been scaled for computational ease. Standard errors are clustered at the focal product level.}} \\
		\end{tabular} 
	}
\end{table} 

Overall, we find that both the scores are significant and predictive of the bundle add-to-cart rate. Since the coefficients are on a logit scale, an example might be fruitful here to highlight the effect. Consider a bundle with complementarity score one standard deviation below the mean such as ``Clorox Regular Bleach'' plus ``Method Gel Hand Soap, Lavender". Additionally, consider a bundle with complementarity score one standard deviation above the mean such as ``Feline Pine Natural Pine Litter Original, Non-Clumping'' plus ``Orijen Cat \& Kitten Biologically Appropriate Grain-Free Chicken, Turkey \& Fish Dry Cat Food''. The mean predicted probability from the hierarchical logistic regression (model 4) of moving from the first bundle to the second bundle increases by 70\%. Analogously, take a bundle with substitutability score one standard deviation below the mean such as ``Welch's Concord Grape Jelly'' plus ``Ragu Cheese Creations, Classic Alfredo" and compare it with a a bundle with score one standard deviation above the mean such as ``Miso-Cup Organic Traditional Soup with Tofu, Single-Serve Envelopes'' plus ``Annie Chun's Rice Express Sticky White Rice, Microwavable Bowls''. The mean predicted probability now increases by 35\%.

\subsection{Across-category robustness}

\begin{figure}[h]
	\centering
	\includegraphics[scale=0.65]{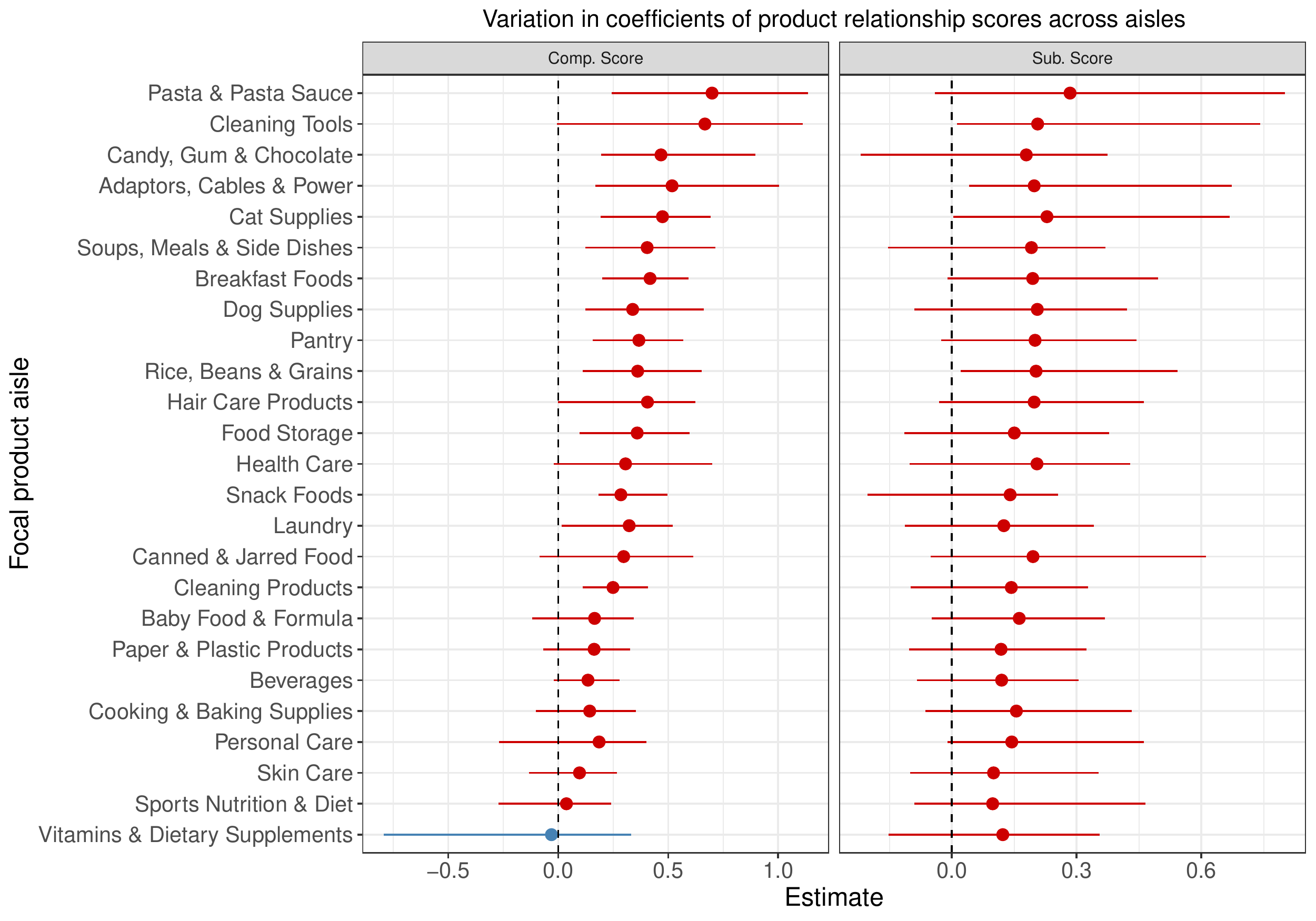}
	\caption{Aisle-varying slopes for product relationship scores from hierarchical logistic regression}
	\label{f:lme_hetero_map}
\end{figure}

We use the varying-slopes in the hierarchical model to examine robustness (homogeneity) of the predictive relationship between complementarity and substitutability scores and add-to-cart across different product aisles. Figure~\ref{f:lme_hetero_map} shows the point estimate of the aisle-specific slopes (including the common slope parameter from Table \ref{t:reg_both_re} ). The plot also includes the $95\%$ confidence intervals generated using a bootstrap clustered on focal product. We plot the varying-slopes for both the heuristics across different product aisles and find that the positive association of complementarity score is fairly robust across all aisles and a higher score is predictive of bundle add-to-cart rate. The point estimates for substitutability score are also positive, however, the confidence intervals for these aisle-specific coefficients are much wider. This is because there is much less variability in $S_{k}$ ($\hat{\sigma}_{\beta^s_k} = 0.06$) as compared to $C_k$ ($\hat{\sigma}_{\beta^b_k} = 0.21$). 

\subsection{Cross-category bundles}

Our hierarchical modeling set-up allows us to learn bundle success likelihood across all product category combinations, including the ones that were not part of our training data. We use the model to generalize our findings outside of our experiment. We randomly create 20,000 out-of-sample bundles from the retailers assortment across all product categories and generate predictions using the model. We aggregate the predictions to category-combination level using the categories of both the products and inspect the patterns we see. 

\begin{figure}[h]
	\centering
	\includegraphics[scale=0.65]{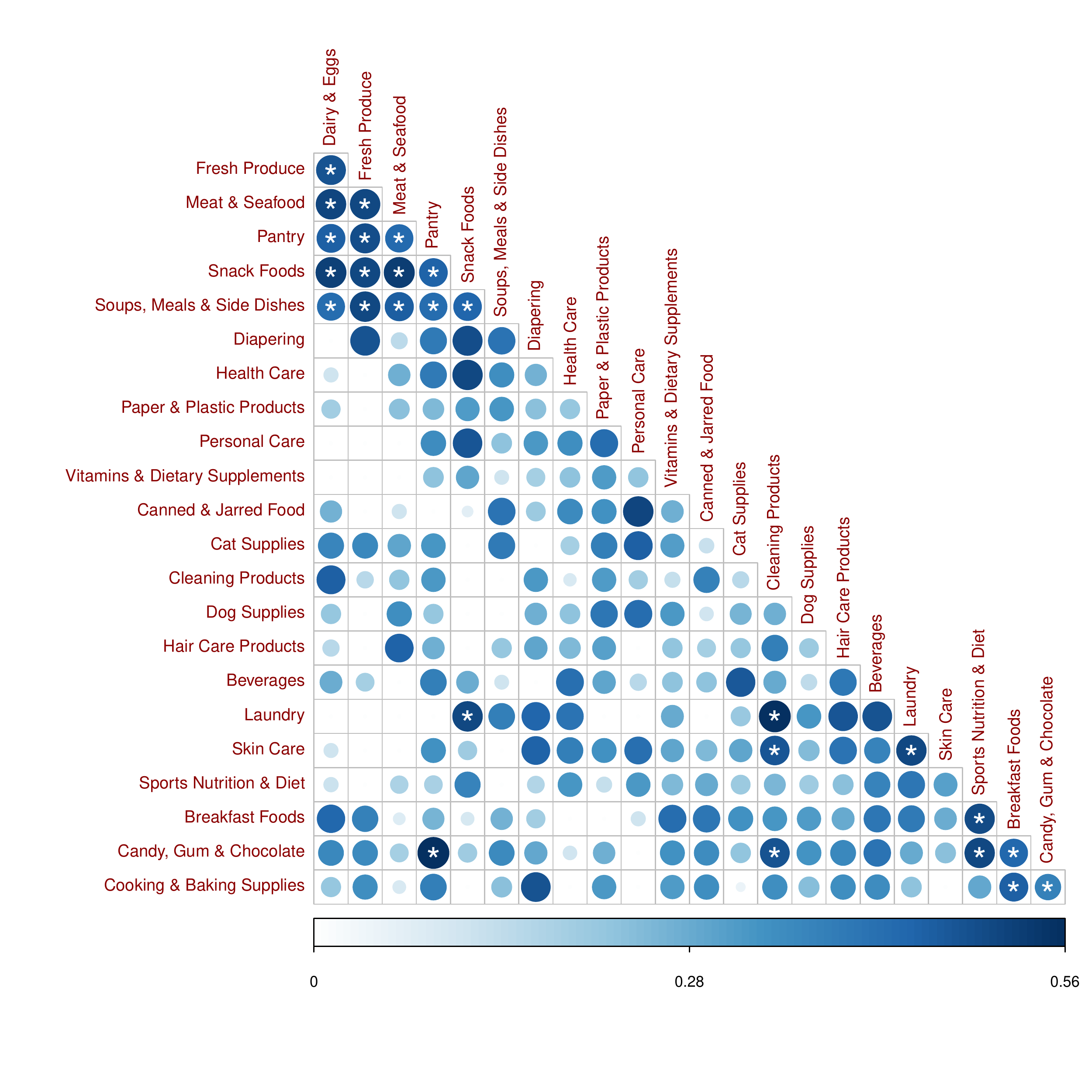}
	\caption{Predicted probabilities aggregated to the aisle combinations.  \normalfont \textit{Note:} The color bar shows \% likelihood of success. (*) Cells highlighted for discussion.}
	\label{f:corr_plot}
\end{figure}

A condensed view of the result is shown in Figure~\ref{f:corr_plot}, which plots the average predicted probability, expressed in percentage, for each aisle combination. Larger darker circles imply higher average likelihood of bundle add-to-cart and the color bar below shows the percent likelihood of success. The product aisle combinations are sorted using spectral clustering. The probabilities are symmetric since we consider each aisle both as the focal product aisle and as the add-on product aisle and then average these. A few interesting patterns are visible and we highlight certain cells for discussion using (*). For example, the two clusters at the extreme ends of the graph --- the top left, and the bottom right, show aisles of products that would be good contenders for cross-category bundles. Fresh produce, dairy and eggs, meat and seafood, snacks, pantry, and soups and side dishes make good bundles with each other. Similarly, sports nutrition products, breakfast foods, and candy make good bundles with each other. Among other combinations, cleaning products go well with laundry, skin care, and interestingly, candy. Candy and chocolates also make a good combination with pantry goods. We also see product combinations in the mid-left of the graph that show cases where cross-category bundling may not be effective. 

In addition to these aisles, we also learn about ``Fresh Produce'', a aisle that was not part of our experiment due to institutional constraints since fresh produce was not available in all the markets. The model still allows us to predict success likelihood for bundles where one or both of the products are from this aisle. The predictions in the case are generated from the grand mean taken from all the aisles that are part of the training data. We note that these results need to be interpreted with caution since they are expected to have higher variability as compared to the aisles shown in Figure~\ref{f:lme_hetero_map}.

From the heatmap, we see that fresh produce is likely to make good bundles with other food categories such as snacks, pantry, and meat and seafood. Further, since our predicted probabilities are actually at the bundle or product-combination level, we can identify exact products which are likely to make good bundles, even though none of these products or even their category were part of the experiment. For example, the top bundles where the focal product is from the fresh produce category are: ``Organic Girl 50/50 Mix, 5 Oz'' + ``Organic Valley, Roast Chicken Breast, Sliced, 6oz'',  ``Organic Cucumber, 1 Ct'' +  ``Organic Creamery Crumbled Cheese, Feta'', and ``Red Beefsteak Tomatoes, 2 Ct'' + ``BelGioioso Parmesan Shredded Cheese, 5 Oz''.

\subsection{Zero co-purchase bundles}

The results from the model in Table~\ref{t:reg_both_re} show that using the embeddings learned in the source problem of learning complements and substitutes are strong predictors of bundle add-to-cart likelihood --- even after controlling for co-purchase rates. Removing the product relationship scores from either model 2 or model 3 in Table~\ref{t:reg_both_re} results in significantly worse fit ($p\text{s} < 0.001$; see Tables~\ref{t:anova_m2} and \ref{t:anova_m3} in the Appendix), highlighting the importance of using the scores. Here we look at a few examples of bundles with high likelihood of success but with no historical co-purchase data. 

\begin{figure}[h]
	\centering
	\includegraphics[scale=0.55]{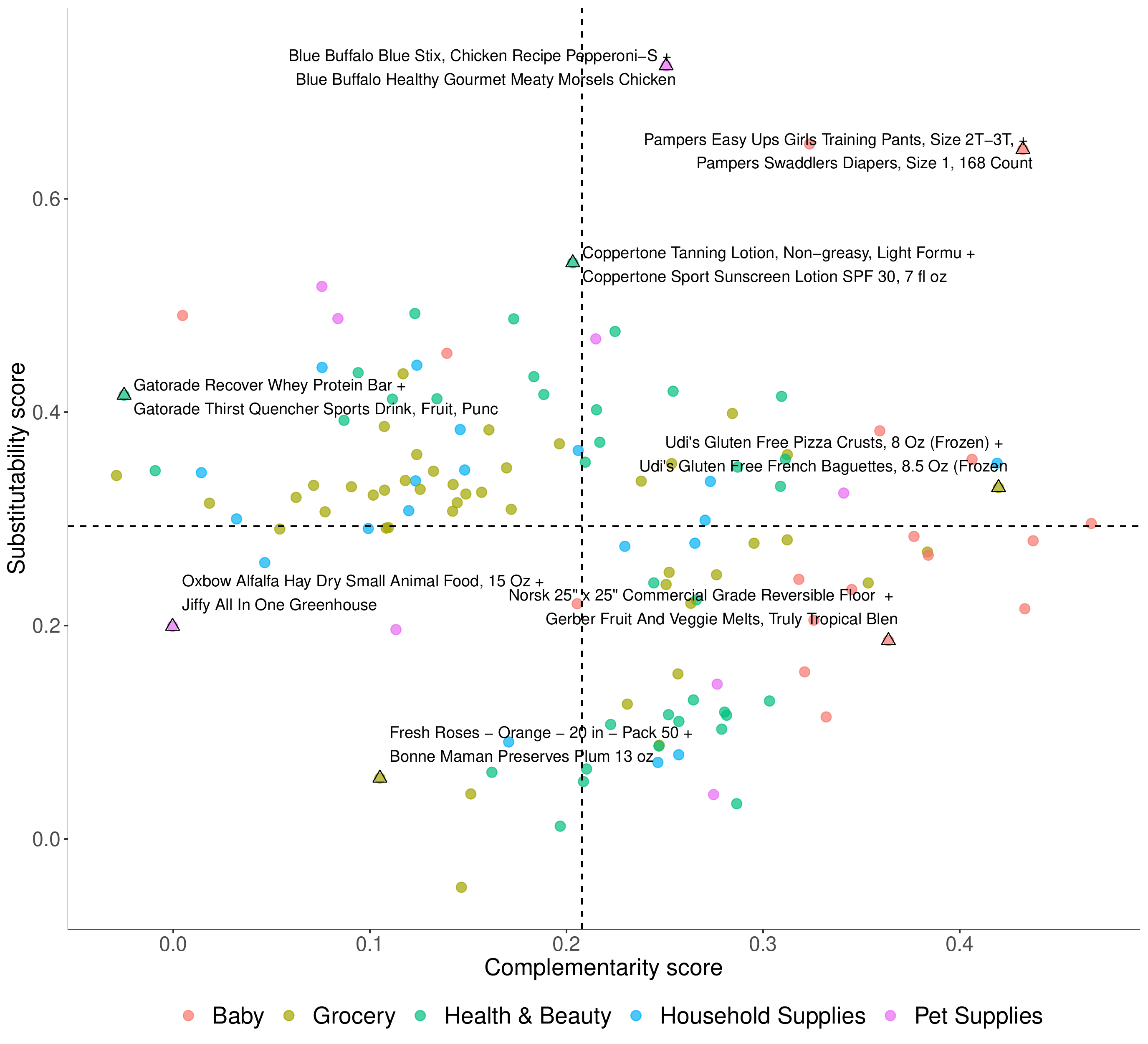}
	\caption{Top 3 predicted bundles from each product aisle with zero co-purchases.   \normalfont \textit{Note:} Dashed lines are the means of the respective scores.}
	\label{f:zero-copur}
\end{figure}

As noted earlier and visually depicted in Figure~\ref{f:copur_comp}, a major benefit of using the embeddings is that we can learn relationships among products that have never been purchased before. Our transfer learning framework can use the embeddings downstream to identify successful bundles even with the zero historical co-purchases. We use the hierarchical logistic regression model to predict the success likelihood of randomly selected 20,000 bundles. Figure~\ref{f:zero-copur} shows a scatter plot of the predicted bundles plotted against the complementarity and substitutability scores. We plot the top-3 predicted bundles for each product aisle of the focal product. We show the product department for clarity and also highlight a few example bundles. The dashed lines are the means of the respective scores. We see from the figure that most of these top-scoring bundles lie on the off-diagonal axis. These bundles have either higher than average complementarity score (lower right quadrant) or higher than average substitutability score (upper left quadrant), which implies that the relationship scores are driving these predictions. We provide more examples of such zero co-purchase bundles in the Appendix in Table~\ref{t:zero_copur_top} where we present the top scoring bundle within each product aisle of the focal category along with the predicted probability.   

\section{Discussion} \label{s:disc}

We propose a novel machine-learning-based bundle design methodology for a large assortment of retail products from multiple categories. Our methodology is based on the historical purchases and considerations sets generated by consumers while shopping online. Using historical product purchases and product views, we create two continuous dense representations of products, called product embeddings, in the purchase space and in the search space respectively. We put minimal structure on these embeddings and create heuristics of complementarity and substitutability between products. In essence, we exploit the notion that products that are ``close'' in the product space can be considered potential complements and products that are ``close'' in the search space can be considered potential substitutes.

Using the heuristics for product relationships, we generate multiple bundles for thousands of products across different product categories and run a field experiment in partnership with an online retailer. We especially create bundles across product categories and using imperfect substitutes to learn consumer preferences across different kinds of bundles. Using the product relationship scores to create bundles allows us to go beyond co-purchases and explore the bundle space in a principled way. We test the effectiveness and robustness of these scores using predictive modeling while controlling for pre-treatment variables and product meta-data. We build a hierarchical logistic regression model with varying intercepts and slopes and the focal product category level. We find that both scores are predictive of bundle success as measured by bundle add-to-cart. 

Our entire approach can be conceptualized according to a transfer learning framework, whereby knowledge gained in one task is subsequently applied to a related task. The first task, also called the source task, is typically one in which there is more data or prior knowledge available. In our case, this is the phase where we learn product embeddings using historical purchase and product views data. The second task, also called the focal task, is in which there is less data. For us, this involves identifying which product pairs would make good promotional bundles. We connect the two tasks using a field experiment that allows us to generate labeled training data for the focal task. The framework moves from machine learning model of co-occurrance to a supervised hierarchical model of purchase decisions.

To the best of our knowledge, ours is the first study to leverage historical purchase and search patterns to generate discount bundles at this scale. Our setting of cross-category online retail is also relatively unexplored in marketing and economics. Moreover, previous studies have primarily been theoretical or lab-based and have typically pre-assumed relationships among products to derive their insights. On the other hand, combining a machine learning model with an online field experiment, we provide empirical evidence and generate generalizable insights from a large number of bundles across multiple product categories. For example, we find that beverages, snacks, and laundry products are good contenders for cross-category bundles with most categories. Meat and seafood go quite well with canned food and fresh produce. On the other hand, health care and baby supplies are not good candidates for cross-category bundles. 

Our study has some constraints as well. We duly note that we trade-off ``structure'' for scale and this has its pros and cons. With our method we are able to work with a much larger set of products and explore a combinatorially complex space efficiently. As a result, we don't focus on the micro-foundations of the model or attempt to tie the model to theory. For instance, we do not look at cross-price elasticities to identify complements or substitutes but rather define them in a way that suits our purpose.  

Additionally, although we include price by controlling for it in the regression models, we do not explicitly include it in the experiment. We believe it would be insightful to randomize the discount in the experiment and investigate the impact on the results. For example, we hypothesize that the retailer would need to provide a smaller discount for complementary bundles and a relatively larger one for variety products. We could not include it in our experiment due to considerations for statistical power but this area is ripe for further exploration.

Another angle which we do not pursue would be to analyze the welfare consequences of bundling in the retail setting. For instance, on one hand bundling products together benefits consumers by reducing search costs, on the other it may be inducing them to make purchases that they would have avoided if not for the discount. Analyzing the net impact on consumer welfare and producer surplus would be interesting to study.



\newpage
\bibliographystyle{informs2014} 
\bibliography{bundling-prod-emb.bib} 



\newpage

\begin{APPENDICES}
\section{Supplementary tables and figures} \label{appendix:app_tab}

\setcounter{table}{0} 
\setcounter{figure}{0} 
\renewcommand{\thetable}{A\arabic{table}} 
\renewcommand{\thefigure}{A\arabic{figure}} 


\begin{table}[h]
	\centering
	\caption{Observation counts from our working sample}
	\scalebox{0.9}{	
		\begin{tabular}{lr}
			\toprule
			{} &    Count \\
			\midrule
			Total users            &  534,284 \\
			Total sessions         &  947,955 \\
			Total purchase baskets &  861,963 \\
			Total search baskets  &   589,552 \\
			Unique products        &  35,000 \\
			\bottomrule
			\multicolumn{2}{p{7cm}}{\SingleSpacedXI \footnotesize{\textit{Note 1:} The table shows the size of our working sample after filtering out purchases and searches involving right tail products. We retain the top-35,000 products that include more than 90\% of the purchases in our sample period. }} \\
			\multicolumn{2}{p{7cm}}{\SingleSpacedXI \footnotesize{\textit{Note 2:} Purchase baskets include products purchased and search baskets include products searched but \textit{not} purchased. The number of searches baskets are less than the number of purchase baskets because we define a product searched only if the user opens the description page of the product. The user can, however, purchase without opening the product description page by directly adding the product to the cart while browsing. }} \\
		\end{tabular}
	}
	\label{t:high_level}
\end{table}

\begin{table}[H]
	\centering
	\caption{Products close to ``Organic Russet Potatoes, 5 Lb (10-12 Ct)" in the purchase space}
	\scalebox{0.9}{	
		\begin{tabular}{lr}
			\toprule
			Product &  Score \\
			\midrule
			Organic Celery Hearts, 16 Oz & 0.75 \\
			Organic Grape Tomatoes, 1 Pint & 0.74 \\
			Organic Green Bell Peppers, 2 Ct & 0.74 \\
			Organic Carrots, 2 Lb & 0.73 \\
			Organic Cauliflower, 1 Ct & 0.73 \\
			Organic Garlic, 8 Oz & 0.72 \\
			The Farmers Hen Large Organic Eggs, 1 Dozen & 0.70 \\
			Organic Bananas, Minimum 5 Ct & 0.69 \\
			Organic Broccoli Crowns, 2 Ct & 0.69 \\
			Organic Romaine Hearts, 3 Ct & 0.69 \\
			\bottomrule
			\multicolumn{2}{p{10cm}}{\SingleSpacedXI \footnotesize{\textit{Note 1:} The score is measure of proximity in the purchase space and is indicative of complementarity. A higher score implies stronger complementarity. The score is normalized such that the maximum possible value is 1.}} \\
		\end{tabular}
	}
	\label{t:pur_sim_potato}
\end{table}

\begin{table}[H]
	\centering
	\caption{Products close to ``Joy Ultra Dishwashing Liquid, Lemon Scent, 12.6 oz" in the purchase space}
	\scalebox{0.9}{	
		\begin{tabular}{lr}
			\toprule
			Product &  Score \\
			\midrule
			Bounty Paper Towels, White, 12 Super Rolls & 0.50 \\
			Tide PODS Plus Downy HE Turbo Laundry Detergent Pacs, April Fresh, 54 count & 0.48 \\
			P\&G 45Oz Cmp Gel Detergent & 0.48 \\
			The Art of Shaving Shave Cream, Sandalwood, 5 Oz & 0.47 \\
			Bounty Towel, Bounty Essentials 12pk & 0.46 \\
			Bounty Paper Towels,  Select-A-Size, 6 Triple Rolls & 0.46 \\
			Lillian Dinnerware Pebbled Plastic Plate, 10.25", Clear, 10 Ct & 0.45 \\
			Saratoga Spring Water, 28 Fl Oz & 0.45 \\
			CLR Stainless Steel Cleaner, 12 Oz & 0.45 \\
			Pepcid Complete Dual Action Acid Reducer and Antacid Chewcap, Berry, 100 Ct & 0.45 \\
			\bottomrule
			\multicolumn{2}{p{16cm}}{\SingleSpacedXI \footnotesize{\textit{Note 1:} The score is measure of proximity in the purchase space and is indicative of complementarity. A higher score implies stronger complementarity. The score is normalized such that the maximum possible value is 1.}} \\
		\end{tabular}
	}
	\label{t:pur_sim_dish}
\end{table}

\begin{table}[H]
	\centering
	\caption{Products similar to ``Neutrogena Oil-Free Acne Wash Redness Soothing Cream Facial Cleanser, 6 Fl. Oz" in the purchase space}
	\scalebox{0.8}{	
		\begin{tabular}{lr}
			\toprule
			Product &  Score \\
			\midrule
			U By Kotex Barely There Daily Liners, 100 Ct & 0.53 \\
			Aveeno Active Naturals Daily Moisturizing Body Yogurt Body Wash, Vanilla and Oats, 18 Fl. Oz & 0.52 \\
			Palmer's Cocoa Butter Formula Bottom Butter Zinc Oxide Diaper Rash Cream Tube, 4.4 Oz & 0.52 \\
			Neutrogena Oil-Free Acne Wash Redness Soothing Facial Cleanser With Salicylic Acid, 6 Fl. Oz. & 0.51 \\
			Neutrogena Oil-Free Acne Face Wash Pink Grapefruit Foaming Scrub, Salicylic Acid Acne Treatment, 4.2 Fl. Oz. & 0.49 \\
			Maybelline New York Fit Me Matte \& Poreless Foundation, Natural Beige, 1 Fl Oz & 0.48 \\
			Secret Invisible Solid Anti-Perspirant Deodorant, Berry Fresh, 2.6 Oz, 2 Ct & 0.47 \\
			Motrin IB, Ibuprofen, Aches and Pain Relief, 100 Count & 0.47 \\
			Nature's Bounty  Hair, Skin \& Nails Gummies Strawberry 2,500 mcg , 80 Ct & 0.47 \\
			Equate Ibuprofen Pain Reliever/Fever Reducer 200 mg Tablets, 100 Ct & 0.46 \\
			\bottomrule
			\multicolumn{2}{p{20cm}}{\SingleSpacedXI \footnotesize{\textit{Note 1:} The score is measure of proximity in the purchase space and is indicative of complementarity. A higher score implies stronger complementarity. The score is normalized such that the maximum possible value is 1.}} \\
		\end{tabular}
	}
	\label{t:pur_sim_face}
\end{table}

\begin{table}[H]
	\centering
	\caption{Products close to ``Organic Russet Potatoes, 5 Lb (10-12 Ct)" in the search space}
	\scalebox{0.9}{	
		\begin{tabular}{lr}
			\toprule
			Product &  Score \\
			\midrule
			Green Giant Organic Golden Potatoes, 3 Lb & 0.95 \\
			Green Giant Organic Red Potatoes, 3 Lb & 0.95 \\
			Organic Russet Potatoes, 3 Lb & 0.95 \\
			Green Giant Klondike Gourmet Petite Purple-Purple Fleshed Potatoes, 24 Oz & 0.93 \\
			Green Giant Klondike Fingerling Potatoes, 24 Oz & 0.93 \\
			Green Giant Golden Potatoes, 5 Lb & 0.92 \\
			Organic Sweet Potatoes, 3 Lb & 0.92 \\
			Green Giant Klondike Petite Red-White Fleshed Potatoes, 24 Oz & 0.92 \\
			The Little Potato Garlic Herb Potato Microwave Kit, 16 Oz & 0.92 \\
			The Little Potato Company Garlic Herb Oven Griller Kit, 16 Oz & 0.91 \\
			\bottomrule
			\multicolumn{2}{p{15cm}}{\SingleSpacedXI \footnotesize{\textit{Note 1:} The score is measure of proximity in the search space and is indicative of substitutability. A higher score implies stronger substitutability. The score is normalized such that the maximum possible value is 1.}} \\
		\end{tabular}
	}
	\label{t:search_sim_potato}
\end{table}

\begin{table}[H]
	\centering
	\caption{Products close to ``Joy Ultra Dishwashing Liquid, Lemon Scent, 12.6 oz" in the search space}
	\scalebox{0.6}{	
		\begin{tabular}{lr}
			\toprule
			Product &  Score \\
			\midrule
			Joy Dishwashing Liquid, Lemon, 5gal Pail & 0.83 \\
			Joy Dishwashing Liquid 38 oz Bottle & 0.79 \\
			Joy Dishwashing Liquid Lemon Scent 12.6 oz Bottle & 0.71 \\
			Palmolive Ultra Anti-Bacterial Dish Soap, Orange, 56 Oz & 0.70 \\
			Ajax Triple Action Dish Soap, Orange, 12.6 Oz & 0.69 \\
			Palmolive Ultra Dish Soap, Orange, 25 Fl Oz & 0.69 \\
			Biokleen  Natural Dish Liquid, Citrus, 32 Oz, 12 Ct & 0.69 \\
			Palmolive OXY Plus Power Degreaser Dish Soap, 10 Oz & 0.69 \\
			Ajax Super Desgreaser Dish Soap, Lemon, 52 Oz & 0.69 \\
			Ajax Dish Soap, Tropical Lime Twist, 52 Oz & 0.68 \\
			\bottomrule
			\multicolumn{2}{p{15cm}}{\SingleSpacedXI \footnotesize{\textit{Note 1:} The score is measure of proximity in the search space and is indicative of substitutability. A higher score implies stronger substitutability. The score is normalized such that the maximum possible value is 1.}} \\
		\end{tabular}
	}
	\label{t:search_sim_dish}
\end{table}

\begin{table}[H]
	\centering
	\caption{Products similar to ``Neutrogena Oil-Free Acne Wash Redness Soothing Cream Facial Cleanser, 6 Fl. Oz" in the search space}
	\scalebox{0.6}{	
		\begin{tabular}{lr}
			\toprule
			Product &  Score \\
			\midrule
			Neutrogena Oil-Free Acne Face Wash With Salicylic Acid, 6 Oz. & 0.84 \\
			Neutrogena Oil-Free Acne Face Wash Daily Scrub With Salicylic Acid, 4.2 Fl. Oz. & 0.84 \\
			Neutrogena Oil-Free Acne Face Wash Pink Grapefruit Foaming Scrub, Salicylic Acid Acne Treatment, 4.2 Fl. Oz. & 0.83 \\
			Neutrogena Naturals Purifying Pore Scrub, 4 Fl. Oz. & 0.82 \\
			Neutrogena Rapid Clear Stubborn Acne Cleanser, 5 Oz & 0.82 \\
			Neutrogena All-In-1 Acne Control Daily Scrub, Acne Treatment 4.2 Fl. Oz. & 0.82 \\
			Neutrogena Oil-Free Acne Wash Pink Grapefruit Cream Cleanser, 6 Oz & 0.82 \\
			Neutrogena Oil-Free Acne Face Wash With Salicylic Acid, 9.1 Oz. & 0.81 \\
			Neutrogena Men Oil-Free Invigorating Foaming Face Wash, 5.1 Fl. Oz & 0.80 \\
			Neutrogena Oil-Free Acne Face Wash Pink Grapefruit Foaming Scrub, Salicylic Acid Acne Treatment, 6.7 Fl. Oz. & 0.80 \\
			\bottomrule
			\multicolumn{2}{p{20cm}}{\SingleSpacedXI \footnotesize{\textit{Note 1:} The score is measure of proximity in the search space and is indicative of substitutability. A higher score implies stronger substitutability. The score is normalized such that the maximum possible value is 1.}} \\
		\end{tabular}
	}
	\label{t:search_sim_face}
\end{table}

		
\begin{table}[h] \centering 
	\caption{Bundle types based on relationship heuristics for complete cases} 
	\label{t:bun_cat_complete} 
	\scalebox{0.9}{	
		\begin{tabular}{@{\extracolsep{5pt}} lrrrr} 
			\\[-1.8ex]\hline 
			\hline \\[-1.8ex]
			&  Co-purchase &  Cross  &   Cross  & Variety \\ 	
			&   &  category &   department  &  \\ 	
			\cmidrule{2-5} \\
			&  (CP) &  (CC) &    (DC) & (VR) \\ 	
			\hline \\[-1.8ex] 
			Count & $747$ & $747$ & $747$ & $747$ \\ 
			Comp. score & $0.43$ & $0.36$ & $0.23$ & $0.42$ \\ 
			Sub. score & $0.61$ & $0.40$ & $0.20$ & $0.75$ \\ 
			Co-purchase rate & $0.56$ & $0.22$ & $0.06$ & $0.45$ \\ 
			Price-1 & $9.49$ & $9.49$ & $9.49$ & $9.49$ \\ 
			Price-2 & $8.62$ & $8.11$ & $11.54$ & $9.84$ \\ 
			Purchase rate-1 & $0.04$ & $0.04$ & $0.04$ & $0.04$ \\ 
			Purchase rate-2 & $0.04$ & $0.05$ & $0.06$ & $0.04$ \\ 
			Product rating-1 & $4.76$ & $4.76$ & $4.76$ & $4.76$ \\ 
			Product rating-2 & $4.76$ & $4.72$ & $4.82$ & $4.75$ \\ 
			Same brand & $0.50$ & $0.20$ & $0.12$ & $0.49$ \\ 
			Same aisle & $0.74$ & $0.58$ & $0$ & $1$ \\ 
			Same category & $0.55$ & $0.01$ & $0$ & $0.98$ \\ 
			\hline \\[-1.8ex] 
			\multicolumn{5}{p{11.5cm}}{\SingleSpacedXI \footnotesize{\textit{Note 1:} Co-purchase rate has been multiplied by 100. Price-1, Purchase rate-1, and Product rating-1 show the average price, average historical purchase rate, and the average product rating of the focal product in each bundle type. Price-2, Purchase rate-2, and Product rating-2 are corresponding variables for the add-on product. Same brand, Same aisle, and Same category are binary variables that indicate if the two products are from the same brand, same aisle, and the same category respectively.}} \\
		\end{tabular} 
	}
\end{table}


	\begin{table}[h] \centering 
		\caption{Pre-experiment summary statistics for key variables} 
		\subcaption*{\tiny Variable means and standard deviations}
		\label{t:pre_exp_norm}
		\scalebox{0.9}{	
			\begin{tabular}{@{\extracolsep{5pt}} lrrrrr} 
				\\[-1.8ex]\hline 
				\hline \\[-1.8ex]
			    &  Co-purchase &  Cross-cat.  &   Cross-dept.  & Variety & p-value \\ 	
			    \cmidrule{2-5} \\
			    &    (CP)           &  (CC)     & (DC)  & (VR) \\ 
				\hline \\[-1.8ex] 
				Observations & 5,722 & 5,451 & 4,925 & 5,474 &  \\ 
				\midrule
				Visits & 3.5 & 3.69 & 3.55 & 3.63 & 0.15 \\ 
				& 5.39 & 5.78 & 5.6 & 5.89 & - \\ 
				Product views & 13.07 & 13.96 & 13.39 & 13.78 & 0.12 \\ 
				& 31.39 & 33.39 & 34.24 & 34.66 & - \\ 
				Products ATC & 0.39 & 0.38 & 0.38 & 0.38 & 0.65 \\ 
				& 0.81 & 0.7 & 0.8 & 0.76 & - \\ 
				Units purchased & 0.56 & 0.52 & 0.53 & 0.54 & 0.62 \\ 
				& 2.19 & 2 & 1.83 & 2.05 & - \\ 
				Revenue & 5.4 & 5.22 & 5.16 & 5.14 & 0.8 \\ 
				& 23.46 & 24.1 & 28.47 & 22.85 & - \\ 
				\hline \\[-1.8ex] 
				\multicolumn{6}{p{13cm}}{\SingleSpacedXI \footnotesize{\textit{Note 1:} Pre-experiment means and standard deviations for users who visited the retailer's website at least once during the month prior to the start of the experiment.}} \\
				\multicolumn{6}{p{13cm}}{\SingleSpacedXI \footnotesize{\textit{Note 2:} Product views include all page visits by the user. Products ATC accounts for the total number of different products added-to-cart whereas units purchased account for multiple units of the same product bought.}} \\
				\multicolumn{6}{p{13cm}}{\SingleSpacedXI \footnotesize{\textit{Note 3:} P-values are from a test of equality of means (ANOVA).}}
			\end{tabular} 
		}
	\end{table}

\begin{table}[h] \centering 
	\caption{Experiment results split by bundle type for products with all four bundle types} 
	\label{t:bp_metrics_complete} 
	\scalebox{0.9}{
		\begin{tabular}{@{\extracolsep{5pt}} lrrrr} 
			\\[-1.8ex]\hline 
			\hline \\[-1.8ex] 
			& CP & CC & DC & VR \\ 
			\hline \\[-1.8ex] 
			Bundles & $747$ & $747$ & $747$ & $747$ \\ 
			Views & $36,231$ & $28,565$ & $28,658$ & $27,769$ \\ 
			Clicks & $599$ & $328$ & $275$ & $499$ \\ 
			ATC & $454$ & $252$ & $103$ & $296$ \\ 
			Purchases & $86$ & $31$ & $6$ & $45$ \\ 
			\hdashline \\
			CTR & $0.016$ & $0.012$ & $0.010$ & $0.018$ \\ 
			ATC rate & $0.012$ & $0.009$ & $0.004$ & $0.011$ \\ 
			Purchase rate & $0.002$ & $0.001$ & $0.0002$ & $0.002$ \\ 
			\hline \\[-1.8ex] 
			\multicolumn{5}{p{8.5cm}}{\SingleSpacedXI \footnotesize{\textit{Note 1:} CTR is click-through rate. ATC is add-to-cart. The rate columns in the right half of the table are calculated as a proportion of views.}} \\
		\end{tabular} 
	}
\end{table} 

\begin{table}[h] \centering 
	\caption{P-values from pairwise proportions tests for bundle add-to-cart and purchase rates} 
	\label{t:prop_test} 
	\scalebox{0.9}{
		\begin{tabular}{@{\extracolsep{5pt}} lrr} 
			\\[-1.8ex]\hline 
			\hline \\[-1.8ex] 
			& ATC & Purchase \\ 
			& Rate & Rate \\ 
			\hline \\[-1.8ex] 
			CP-CC & $<0.001$ & $<0.001$ \\ 
			CP-DC & $<0.001$ & $<0.001$ \\ 
			CP-VR & $<0.001$ & $0.054$ \\ 
			\hdashline
			CC-DC & $<0.001$ & $<0.001$ \\ 
			CC-VR & $<0.001$ & $0.003$ \\ 
			\hdashline
			DC-VR & $<0.001$ & $<0.001$ \\ 
			\hline \\[-1.8ex] 
			\multicolumn{3}{p{5cm}}{\SingleSpacedXI \footnotesize{\textit{Note:} Pairwise tests compare the add-to-cart rate and the bundle purchase rate across different bundle types. }} \\
		\end{tabular} 			
	}
\end{table} 

\begin{sidewaystable}[h]
\centering
\caption{Top-3 performing bundles from each experiment arm} 
  \label{t:top_3_btype} 
  \scalebox{0.6}{
        \begin{tabular}{@{\extracolsep{5pt}} lllllc} 
        \\[-1.8ex]\hline 
        \hline \\[-1.8ex] 
        Bundle Type & Category-1 & Product-1 & Category-2 & Product-2 & ATC count \\ 
        \hline \\[-1.8ex] 
        CP & Scent Boosters & Downy Unstopables In-Wash Premium Scent Booster-FRESH & Scent Boosters & Downy Unstopables In-Wash Premium Scent Booster-LUSH & $8$ \\ 
        CP & Mops and Accessories & O'Cedar EasyWring Microfiber Wet Mop and Bucket Syst & Mops and Accessories & Easy Wring Mop Refill & $7$ \\ 
        CP & Chips and Pretzels & Utz Party Mix Barrel, 43 Oz & Popcorn and Puffed Snacks & Utz Cheese Balls Barrel, 35 Oz & $6$ \\ 
        CC & Scent Boosters & Downy Unstopables In-Wash Premium Scent Booster-FRESH & Dryer Sheets & Downy Infusions Botanical Mist Fabric Softener Dry & $10$ \\ 
        CC & Coffee & Custom Variety Pack Single Serve for Keurig, 40 Ct & Cream and Creamers & International Delight Coffeehouse Inspirations Sin & $5$ \\ 
        CC & Chocolate & Mars  Chocolate Favorites Mini Bars Variety Mix Ba & Chewy Candy & Skittles/Lifesavers/Starburst Candy Variety Pack,  & $4$ \\ 
        DC & Protein and Meal Replacement & Premier Protein High Protein Shake, Chocolate, 11  & Protein and Granola Bars & Pure Protein Bar, Chocolate Peanut Butter, 1.76 Oz & $3$ \\ 
        DC & Laundry Detergent & GreenShield Organic  Laundry Detergent, Free and Cle & Hand Soap & ECOS Hand Soap Refill, Lavender, 32 Fl Oz & $3$ \\ 
        DC & Baking Ingredients & Arm and Hammer Pure Baking Soda, 3.5 lbs & Other Laundry Care & 20 Mule Team Borax Detergent Booster, 76 Ounces & $3$ \\ 
        VR & Pasta and Noodles & Barilla Gluten Free Rotini, 12 Oz & Pasta and Noodles & Barilla Gluten Free Penne, 12 Oz & $5$ \\ 
        VR & Glass Cleaners & Windex Outdoor All In One Glass Cleaninig Set & Glass Cleaners & Windex Outdoor All-In-One Glass Cleaning Kit Refil & $4$ \\ 
        VR & Deodorants & Old Spice Pure Sport Deodorant, 3 Oz, 5 Ct & Deodorants & Degree Dry Protection Deodorant Bonus Pack, Shower & $4$ \\ 
        \hline \\[-1.8ex] 
        \end{tabular} 
}
\end{sidewaystable}

\begin{table}[h] \centering 
  \caption{Summary statistics for variables used in the predictive model}
  \label{t:sum_stat_model} 
  \scalebox{0.9}{
    \begin{tabular}{@{\extracolsep{5pt}} lccccccccc} 
        \\[-1.8ex]\hline 
        \hline \\[-1.8ex] 
         &  &  &  &  &  &  &  & Mean: & Mean: \\
        Variable & Mean & SD & Min & Q25 & Q50 & Q75 & Max & ATC$=$0 & ATC$>$0 \\
        \hline \\[-1.8ex] 
        Comp. score & $0.345$ & $0.173$ & $$-$0.219$ & $0.222$ & $0.340$ & $0.467$ & $0.926$ & $0.331$ & $0.462$ \\ 
        Sub. score & $0.488$ & $0.283$ & $$-$0.212$ & $0.238$ & $0.481$ & $0.749$ & $0.984$ & $0.471$ & $0.626$ \\ 
        Hist. co-purchase rate & $0.003$ & $0.009$ & $0.000$ & $0.000$ & $0.001$ & $0.003$ & $0.254$ & $0.003$ & $0.006$ \\
        Price-1 & $9.470$ & $12.811$ & $1.750$ & $3.770$ & $5.490$ & $9.980$ & $281$ & $9.510$ & $9.138$ \\ 
        Price-2 & $8.947$ & $11.776$ & $1.750$ & $3.780$ & $5.180$ & $9.470$ & $182.550$ & $9.096$ & $7.712$ \\ 
        Rel. price interaction & $0.000$ & $1.000$ & $$-$17.834$ & $$-$0.017$ & $$-$0.016$ & $$-$0.013$ & $42.874$ & $$-$0.005$ & $0.039$ \\ 
        Hist. Purchase Rate-1 & $0.039$ & $0.026$ & $0.002$ & $0.023$ & $0.032$ & $0.047$ & $0.516$ & $0.038$ & $0.041$ \\ 
        Hist. purchase rate-2 & $0.049$ & $0.033$ & $0.001$ & $0.027$ & $0.038$ & $0.059$ & $0.287$ & $0.049$ & $0.048$ \\ 
        Rating-1 & $4.708$ & $0.415$ & $0$ & $4.600$ & $4.800$ & $5$ & $5$ & $4.703$ & $4.747$ \\ 
        Rating-2 & $4.732$ & $0.390$ & $0$ & $4.600$ & $4.900$ & $5$ & $5$ & $4.732$ & $4.730$ \\ 
        Same brand & $0.301$ & $0.459$ & $0$ & $0$ & $0$ & $1$ & $1$ & $0.277$ & $0.499$ \\ 
        Same category & $0.401$ & $0.490$ & $0$ & $0$ & $0$ & $1$ & $1$ & $0.385$ & $0.531$ \\ 
        Diff. category \textbar  Same aisle & $0.196$ & $0.397$ & $0$ & $0$ & $0$ & $0$ & $1$ & $0.192$ & $0.233$ \\ 
        \hline \\[-1.8ex]
        \hline 
        \hline \\[-1.8ex] 
        \multicolumn{10}{p{18cm}}{\SingleSpacedXI \footnotesize{\textit{Note:} Relative price interaction is the signed and squared difference between the prices of the two products. The last two columns show the mean of each variable for bundles that were never added-to-cart and those that were added-to-cart at least once.}} \\
    \end{tabular} 
}
\end{table} 

\begin{table}[h] \centering 
  \caption{ANOVA for Model 2 with and without product relationship scores} 
  \label{t:anova_m2} 
  \scalebox{0.9}{
    \begin{tabular}{@{\extracolsep{5pt}} ccccc} 
    \\[-1.8ex]\hline 
    \hline \\[-1.8ex] 
     & Resid. Dev & Df & Deviance & Pr(\textgreater Chi) \\ 
    \hline \\[-1.8ex] 
    Model 2 - without scores & $4,388.53$ & $-$ & $-$ & $-$ \\ 
    Model 2 - with scores & $4,317.03$ & $2$ & $71.50$ & $<0.001$ \\ 
    \hline \\[-1.8ex] 
    \end{tabular} 
    }
\end{table} 

\begin{table}[h] \centering 
  \caption{ANOVA for Model 3 with and without product relationship scores} 
  \label{t:anova_m3} 
  \scalebox{0.9}{
        \begin{tabular}{@{\extracolsep{5pt}} ccccc} 
        \\[-1.8ex]\hline 
        \hline \\[-1.8ex] 
         & Deviance & Chisq & Chi Df & Pr(\textgreater Chisq) \\ 
        \hline \\[-1.8ex] 
        Model 3 - without scores & $6,620.75$ & $-$ & $-$ & $-$ \\ 
        Model 3 - with scores & $6,551.41$ & $69.34$ & $2$ & $<0.001$ \\ 
        \hline \\[-1.8ex] 
        \end{tabular} 
        }
\end{table} 

\begin{table}[h] \centering 
  \caption{Top predicted cross-category bundles with zero historical co-purchases} 
  \label{t:zero_copur_top} 
  \scalebox{0.5}{
\begin{tabular}{@{\extracolsep{5pt}} llllr} 
\\[-1.8ex]\hline 
\hline \\[-1.8ex] 
Category-1 & Product-1 & Category-2 & Product-2 & Pred. Prob. \\ 
\hline \\[-1.8ex] 
Cookies & Enjoy Life Foods Gluten Free Soft Baked Mini Cooki & Chips and Pretzels & Enjoy Life Foods Gluten Free Lentil Chips, Garlic  & $0.018$ \\ 
Laundry Detergent & OxiClean Sparkling Fresh Laundry Detergent Paks, 4 & Stain Removers & OxiClean Versatile Stain Remover, 28.32 Ounces & $0.017$ \\ 
Facial Cleansers & Alba Botanica Hawaiian Facial Cleanser, Pineapply  & Sun Care & Alba Botanica Hawaiian Natural Sunblock SPF 45 Rev & $0.016$ \\ 
Cold Cereals & Cascadian Farm Ancient Grain Granola Cereal 95\% Or & Granola and Muesli & Cascadian Farms Organic Granola Cereal French Vani & $0.015$ \\ 
Glass Cleaners & Formula 409 Glass and Surface Cleaner Concentrate, 1 & Surface Care and Protection & Formula 409 Stone and Steel Cleaner, Spray Bottle, & $0.014$ \\ 
Face Makeup & e.l.f. Blush Palette, Dark, 0.56 Oz & Makeup Brushes & e.l.f. Flawless Concealer Brush & $0.013$ \\ 
Macaroni and Cheese & Annie's Deluxe Elbows and Four Cheese Sauce, 10 Oz & Soups & Annie's Homegrown Organic Soup, Tomato, 17 Oz & $0.013$ \\ 
Turkey & Hillshire Farm Turkey Sausage, 13 Oz & Pork & Hillshire Farm Hot Links Smoked Sausage, 0.875 Lb & $0.013$ \\ 
Shampoos & Garnier Whole Blends Hydrating Shampoo, Coconut Wa & Conditioners & Garnier Whole Blends Repairing Conditioner, Honey  & $0.013$ \\ 
Dusting Tools and Cloths & Swiffer Sweeper Dry Cloth Refill, Lavender Vanilla & Mops and Accessories & Swiffer Sweeper with FeBreze lavender vanilla 80 D & $0.013$ \\ 
Chocolate & Nestle TOLL HOUSE Peanut Butter and Milk Chocolate M & Hard Candy and Lollipops & Brachs Premium Select Cane Mix 12ct & $0.013$ \\ 
Fresh Juice and Fruit Drinks & Califia Farms Tangerine Juice, 48 Fl Oz & Non-Dairy Milks and Creamers & Califia Farms Almondmilk Creamer, Hazelnut, 25.4 F & $0.012$ \\ 
Food Storage Bags & Glad Zipper Food Storage Bags, Snack Size, 50 Ct & Food Storage Containers & Glad Big Bowl Food Storage Containers, Round, 48 O & $0.011$ \\ 
Rice & Near East Toasted Almond Rice Pilaf Mix, 6.6 Ounce & Couscous & Near East Wild Mushroom Herb Couscous 5.4 Ounce & $0.011$ \\ 
Pasta Sauces & Barilla All Natural Pasta Sauce, Spicy Marinara, 2 & Pasta and Noodles & Barilla Fideo Cut Spaghetti, 16 Oz & $0.011$ \\ 
Condiments & Walden Farms Ranch Mayo Jar 12 Ounce By Walden Far & Salad Dressings & Walden Farms Ranch Salad Dressing, 12 Oz & $0.011$ \\ 
Shaving Cream & Gillette Satin Care Sensitive Skin Shave Gel for W & Razors & Gillette Fusion Gift Set 5 pc Box & $0.010$ \\ 
Vegetables & White Asparagus, 1 Bunch & Fruit & Giorgio Portabella Mushroom Caps, 6 Oz & $0.010$ \\ 
Paper Napkins & Seventh Generation Lunch Napkins, White, 250 Ct & Disposable Tableware & Dixie Plastic Tableware Heavy Mediumweight Knives  & $0.010$ \\ 
Dog Food & Blue Buffalo Wilderness Grain-Free Duck Recipe Dry & Dog Treats & Blue Buffalo Blue Stix, Chicken Recipe Pepperoni-S & $0.010$ \\ 
Letter Vitamins & Spring Valley Vitamin E 400 IU Softgels, 100 Ct & Supplements & Spring Valley Odorless Garlic 1000 mg Softgels, 12 & $0.009$ \\ 
Baking Mixes & Rumford Baking Powder - Aluminum Free - Non-Gmo -  & Flours and Meals & Gold Medal Wondra Quick-Mixing All-Purpose Flour,  & $0.009$ \\ 
Fresh Bakery Bread & Asiago Bagels, 4 Oz (Pack of 6) & Packaged Bread & Alvarado St. Bakery Sprouted Wheat Cinnamon Raisin & $0.009$ \\ 
Yogurt & Breakstone's Small Curd Cottage Cheese, 4 X 4 Oz & Salsas and Dips & Breakstone's Sour Cream, Reduced Fat, 16 Oz & $0.008$ \\ 
Baby and Toddler Snacks & Baby Mum-Mum Vegetable Flavor Rice Biscuit, 24-pie & Baby Food & Earth's Best Organic Baby Food, 2nd Vegetables - S & $0.008$ \\ 
Protein and Meal Replacement & Gatorade Recover Protein Shake, Chocolate, 11.16 F & Sports and Energy Drinks & Gatorade Thirst Quencher Sports Drink, Lemon-Lime, & $0.008$ \\ 
Canned Tomatoes and Paste & Roland Sun Dried Tomatoes with Olive Oil, 6.3 Oz & Canned Vegetables & Roland Grilled Artichoke Hearts, 8.3 Oz & $0.008$ \\ 
Cat Treats & Blue Buffalo Kitty Cravings Shrimp Crunchy Cat Tre & Cat Food & Blue Buffalo Wilderness Trail Toppers Grain-Free W & $0.008$ \\ 
Tableware & Vanity Fair Impressions 3-ply White Disposable Tab & Cake Supplies & Wilton Celebrate Standard Cupcake Liners, 150 Ct & $0.008$ \\ 
Bath Sponges and Tools & Aveeno Soothing Bath Treatment For Itchy, Irritate & Soap and Body Wash & Aveeno Skin Relief Gentle Scent Body Wash, Nourish & $0.007$ \\ 
Digestion and Nausea & Equate Anti-Diarrheal, Loperamide 2 mg Caplets, 20 & Allergy, Sinus and Asthma & Equate Allergy Relief Cetirizine Hydrocloride Tabl & $0.006$ \\ 
Candles & Glade 2 in 1 Candle, Moonlit Walk and Wandering Stre & Air Fresheners & Glade Automatic Spray Air Freshener Refill, Apple  & $0.006$ \\ 
Training Pants & Pampers Easy Ups Boys Training Pants, Size 2T-3T,  & Diapers & Pampers Swaddlers Diapers, Size Newborn, 32 Count & $0.004$ \\ 
Nail Polish Remover & Cutex Ultra-Powerful Nail Polish Remover, 3.38 Oz & Hand Sanitizers and Wipes & Medline ReadyBath Total Body Cleansing Standard We & $0.004$ \\ 
Baby Bottles and Accessories & OXO Tot Bottle Brush With Stand - Green & Sun Care & Babyganics Mineral-Based Baby Sunscreen Spray, SPF & $0.004$ \\ 
Play Mats and Activity Gyms & Norsk 25" x 25" Commercial Grade Reversible Floor  & Baby Food & Gerber Multigrain Baby Cereal, 8 oz & $0.003$ \\ 
Food & Oxbow Alfalfa Hay Dry Small Animal Food, 15 Oz & Grow Kits and Seed Starters & Jiffy All In One Greenhouse & $0.003$ \\ 
Baby Gift Sets and Baskets & Seventh Generation Coconut Care Gift Pack, Varies & Laundry Detergent & Attitude Baby Sensitive Skin Care Laundry Detergen & $0.003$ \\ 
Men's Fragrance & Calvin Klein Eternity for Men Eau De Toilette, 3.4 & Shampoos & Axe Dark Temptation 2-in-1 Shampoo and Conditioner,  & $0.002$ \\ 
Baby Grooming & Weleda Baby Derma Body Lotion, White Mallow, 6.8 O & Soap and Body Wash & California Baby Shampoo and Body Wash, Swimmers Defe & $0.002$ \\ 
Fresh Cut Flowers & Fresh Roses - Orange - 20 in - Pack 50 & Jams, Jellies and Preserves & Bonne Maman Preserves Plum 13 oz & $0.002$ \\ 
Mobility Aids and Equipment & Drive Medical Winnie Lite Supreme 3 Wheel Walker R & Chewy Candy & Skittles original bite-size easter candies, 7.3 oz & $0.001$ \\ 
\hline \\[-1.8ex] 
\end{tabular} 
}
\end{table} 


\newpage

\begin{figure}[h!]
	\centering
	\begin{subfigure}{.5\textwidth}
		\centering
		\includegraphics[width=1.0\linewidth]{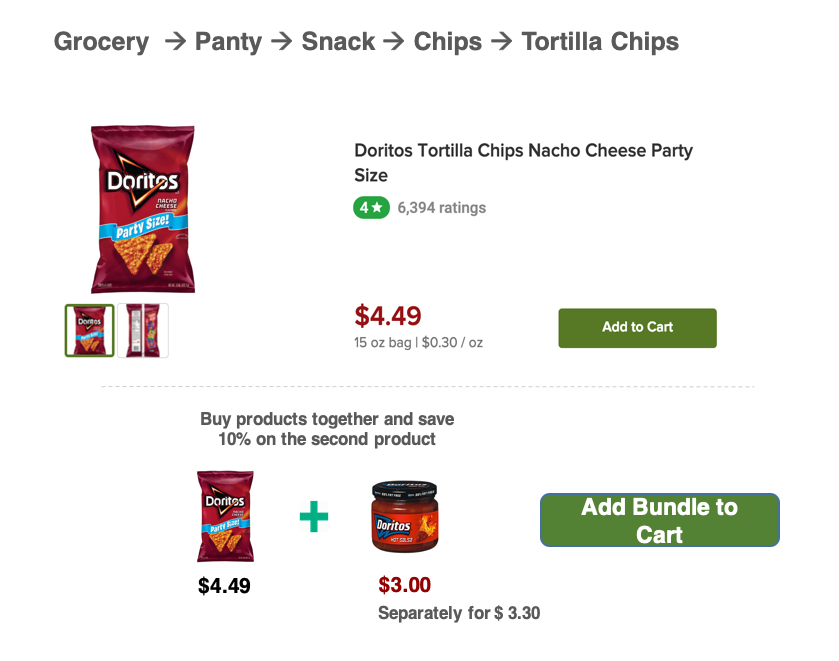}
		\caption{Complementary bundle example}
		\label{f:pilot_comp}
	\end{subfigure}%
	\begin{subfigure}{.5\textwidth}
		\centering
		\includegraphics[width=1.0\linewidth]{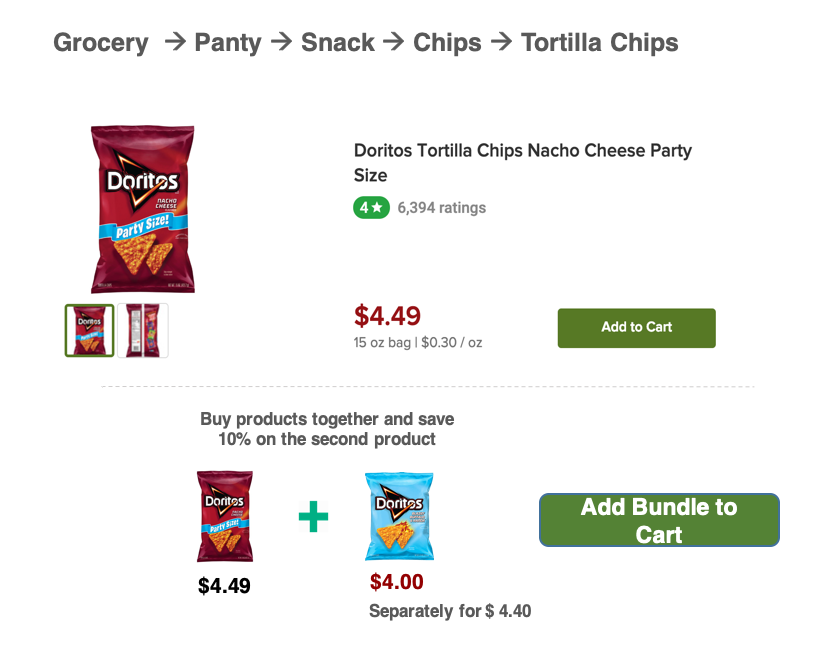}
		\caption{Variety bundle example}
		\label{f:pilot_sub}
	\end{subfigure}
	\caption{Illustrative example from the field experiment}
	\label{f:pilot_example}
\end{figure}

\begin{figure}[h!]
	\centering
	\includegraphics[scale=0.5]{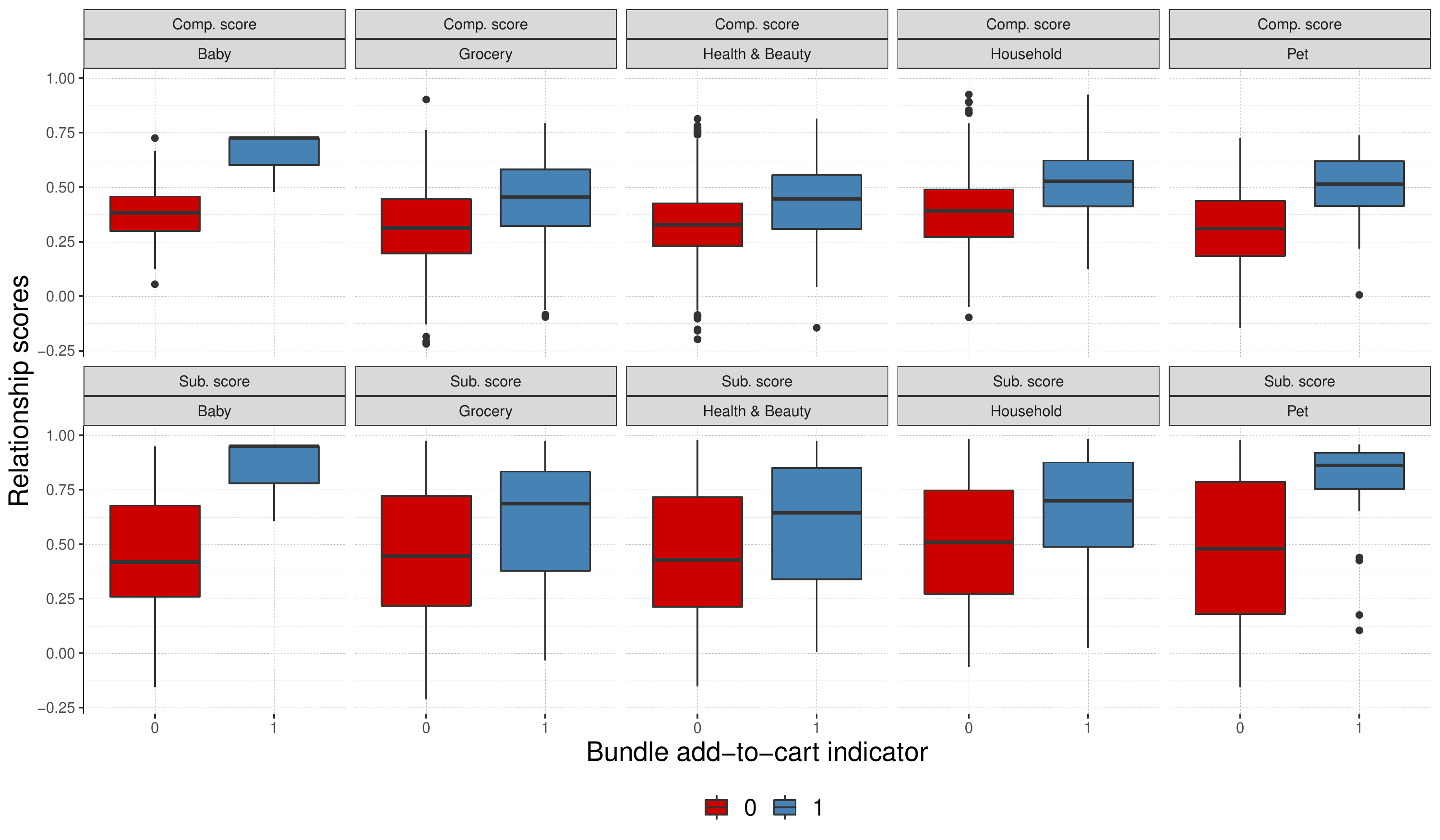}
	\caption{Distribution of product relationship scores across departments split by bundle performance}
	\label{f:box_l0_atc}
\end{figure}

\end{APPENDICES}

\end{document}